\documentclass[sigconf, nonacm]{acmart}

\usepackage{tikz}
\usetikzlibrary{shapes.geometric,calc,positioning,3d,fadings}
\usetikzlibrary{shapes.geometric,positioning,calc,arrows.meta}

\tikzfading[name=shadowfade,
    inner color=transparent!0,
    outer color=transparent!100]

\usepackage{multirow}
\usepackage{booktabs}
\usepackage{graphicx}
\AtBeginDocument{%
  }

\setcopyright{acmlicensed}
\copyrightyear{2027}
\acmYear{2027}
\acmDOI{XXXXXXX.XXXXXXX}
\acmConference[SenSys '27]{The 25th ACM Conference on Embedded Networked Sensor Systems}{May 17--20,
  2027}{Boulder, CO, USA}
\acmISBN{978-1-4503-XXXX-X/2027/06}

\begin{document}
\fancypagestyle{firstpagestyle}{\fancyhf{}\renewcommand{\headrulewidth}{0pt}\fancyhead[C]{\raisebox{8mm}[0pt][0pt]{\parbox{\textwidth}{\centering\footnotesize\textcolor{red}{Preprint. This work has been submitted for possible publication and is currently under review.
}}}}}

\title{RADAR Perception for Dynamic Obstacle Avoidance onboard small-scale Quadrotor UAVs}

\author{Dnyandeep Mandaokar}
\email{dnyandeep.mandaokar@aau.at}
\orcid{0009-0008-5180-8495}
\affiliation{%
  \institution{University of Klagenfurt}
  \city{Klagenfurt}
  \state{Carinthia}
  \country{Austria}
}

\author{Bernhard Rinner}
\affiliation{%
  \institution{University of Klagenfurt}
  \city{Klagenfurt}
  \state{Carinthia}
  \country{Austria}}
\email{bernhard.rinner@aau.at}







\renewcommand{\shortauthors}{Mandaokar et al.}

\begin{abstract}
    Fast dynamic obstacle avoidance (DOA) on uncrewed aerial vehicles (UAVs) demands not only low-latency control and actuation but also reliable, long-range perception to detect obstacles and estimate their motion in time to maneuver. We present, to the best of our knowledge, the first mmWave RADAR-based perception-and-control system for fast onboard DOA. We derive and analyze latency and spatial bounds that relate sensing range, relative speed, and control delay, yielding sufficient conditions for successful avoidance. We further extend prior latency analyses to include the decision latency of the tracker and trigger to evaluate them per trial. Our computationally efficient system combines a lightweight interacting multiple models (IMM) tracker with a control-barrier functions (CBF) controller that directly outputs evasive accelerations. In 300 throw experiments spanning three object sizes and varying visibility (light and dark), it achieves mean position errors of at most 0.15\,m, 0.93\,m, and 0.87\,m in the x, y, and z directions, with a similar spread in 90 experiments in smoke. An onboard implementation on a Raspberry Pi 4B demonstrates real-time feasibility with an end-to-end processing latency of approximately 14\,ms per RADAR frame and a system latency of 75\,ms from first detection to the avoidance command. The UAV successfully avoided obstacles in all 20 tethered indoor flights and all 39 untethered outdoor avoidance maneuvers across three ball sizes, during both hover and forward flights at up to 5\,m/s with measured closing speeds up to 14.6\,m/s.
\end{abstract}





\begin{teaserfigure}
 \includegraphics[width=\textwidth]{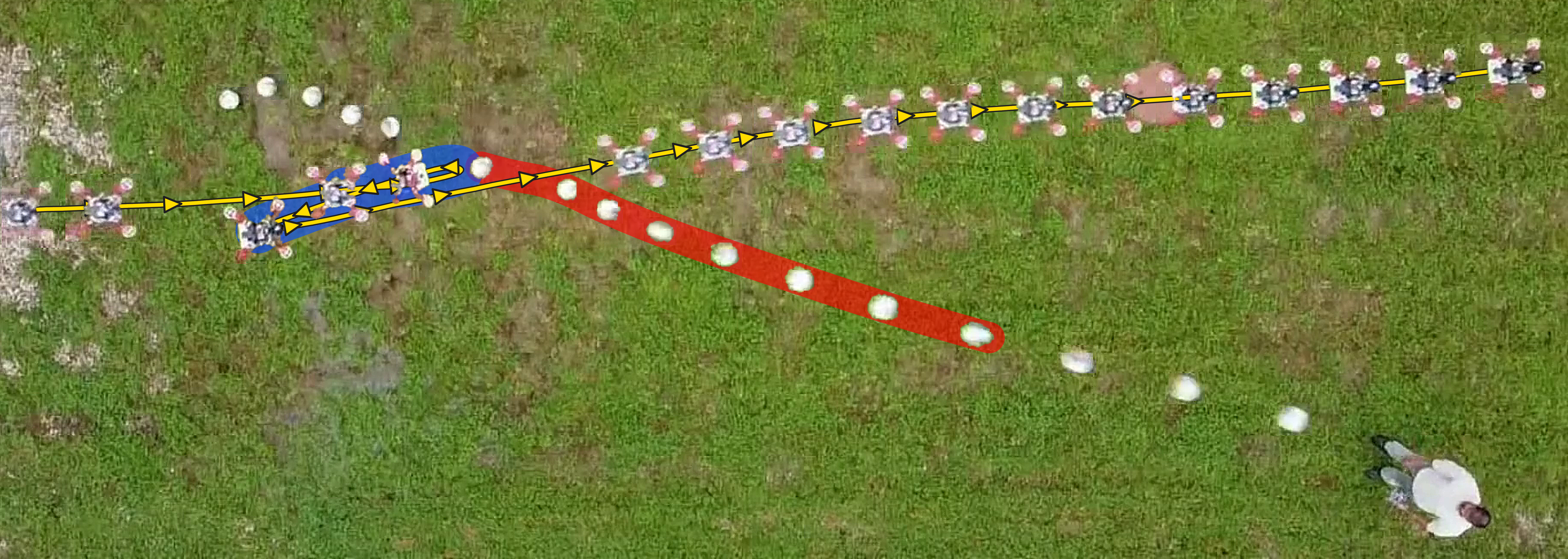}
  \caption{A motion composite image of a quadrotor flying straight at 5~\,m/s and avoiding an approaching ball with mmWave RADAR perception. The red trace illustrates the ball's trajectory from triggering the avoidance maneuver till its execution, and the blue trace represents the quadrotor's avoidance maneuver.}
  \Description{Experiment instance taken from a video to illustrate the avoidance of a flying ball towards the quadrotor. }
  \label{fig:avoidance}
\end{teaserfigure}


\maketitle

\section{Introduction}

Uncrewed aerial vehicles (UAVs) are becoming ubiquitous in applications such as delivery, first aid, and recreational use~\cite{reviewsensinguav2020,Rinner_Computer2021}. These applications require UAVs to operate in a dynamic environment with limited manual control, necessitating safe, autonomous dynamic obstacle avoidance (DOA). Current DOA in UAVs relies heavily on visual perception and control~\cite{rgbdsmalldoa2022, onboardsensinguav2021}. Camera-based perception benefits from small sensor sizes and readily available datasets, which have been used effectively for avoidance in UAVs~\cite {miniOA2023}. However, optical sensing is affected by poor lighting, smoke, and reflective surfaces, often failing in the very environments where robust autonomy is most needed~\cite{doafalanga2020,falangaral2019}. More importantly, optical sensing faces four crucial limitations: detection range, detection latency, velocity estimation of the detected object, and computation load. While event cameras provide low-latency~\cite{rapidoaicra2025,dynamicperception2025}, their poor detection performance in low-contrast conditions limits their use for standalone DOA sensing.
Similarly, light detection and ranging (LIDAR) provides a large detection range, but the LIDAR-based DOA pipelines in~\cite{FAPP2025,pointcloudoa2022,vrba2024onboard} require higher computational resources. Moreover, cameras and LIDARs require additional processing to estimate object speed, thereby increasing latency.

\begin{figure}[t]
  \centering
  \resizebox{\linewidth}{!}{
\begin{tikzpicture}[
    x={(0.6cm, -0.4cm)},  
    y={(1cm, 0.15cm)},    
    z={(0cm, 1cm)},       
    >=stealth
]

    \colorlet{egogreen}{green!60!black}
    \colorlet{obscrit}{orange!90!black}
    \colorlet{fovblue}{cyan!60!blue}
    \colorlet{trajblue}{blue!70!black}
    \colorlet{trackred}{red!85!black}

    \tikzset{
        drone/.pic={
            \begin{scope}[x={(1cm,0cm)}, y={(0cm,1cm)}]
                \draw[line width=1.8pt, black!60, line cap=round] (0,0.02) -- (-0.36,0.16);
                \draw[line width=1.8pt, black!60, line cap=round] (0,0.02) -- ( 0.36,0.16);
                \draw[line width=1.8pt, black!60, line cap=round] (0,-0.02) -- (-0.42,-0.16);
                \draw[line width=1.8pt, black!60, line cap=round] (0,-0.02) -- ( 0.42,-0.16);
                \filldraw[gray!20, draw=black!55, line width=0.9pt, fill opacity=0.85]
                    (-0.36,0.16) ellipse [x radius=0.19cm, y radius=0.065cm];
                \filldraw[gray!20, draw=black!55, line width=0.9pt, fill opacity=0.85]
                    ( 0.36,0.16) ellipse [x radius=0.19cm, y radius=0.065cm];
                \filldraw[gray!20, draw=black!55, line width=0.9pt, fill opacity=0.85]
                    (-0.42,-0.16) ellipse [x radius=0.22cm, y radius=0.075cm];
                \filldraw[gray!20, draw=black!55, line width=0.9pt, fill opacity=0.85]
                    ( 0.42,-0.16) ellipse [x radius=0.22cm, y radius=0.075cm];
                \filldraw[black!80] (0,-0.02) ellipse [x radius=0.17cm, y radius=0.09cm];
                \filldraw[black!55] (0, 0.045) ellipse [x radius=0.13cm, y radius=0.065cm];
            \end{scope}
        }
    }

    \def\shadowZ{-2.0}


    \begin{scope}[canvas is xy plane at z=\shadowZ]
        \fill[white, opacity=0.3] (-4,-2.5) rectangle (4.5,10.5);
        \draw[gray!25, very thin, opacity=0.5, step=1] (-4,-2.5) grid (4.5,10.5);
    \end{scope}

    \begin{scope}[canvas is xy plane at z=\shadowZ]
        \fill[black, path fading=shadowfade, opacity=0.45] (0,0) circle (0.8);
        \fill[black, path fading=shadowfade, opacity=0.15] (0,4.5) circle (0.8);
        \fill[black, path fading=shadowfade, opacity=0.30] (0,6.5) circle (0.5);
        \fill[black, path fading=shadowfade, opacity=0.09] (0,5.35) circle (0.5);
        \fill[black, path fading=shadowfade, opacity=0.30] (0,9.5) circle (0.5);
    \end{scope}



    \coordinate (A)   at (0,0,0);        
    \coordinate (FTL) at (-3, 9,  3);
    \coordinate (FTR) at ( 3, 9,  3);
    \coordinate (FBR) at ( 3, 9, -1.6);
    \coordinate (FBL) at (-3, 9, -1.6);

    \fill[fovblue, fill opacity=0.08] (FTL) -- (FTR) -- (FBR) -- (FBL) -- cycle;
    \fill[fovblue!70!black, fill opacity=0.08] (A) -- (FBL) -- (FBR) -- cycle;
    \fill[fovblue, fill opacity=0.06] (A) -- (FTL) -- (FBL) -- cycle;
    \fill[fovblue, fill opacity=0.06] (A) -- (FTR) -- (FBR) -- cycle;
    \fill[fovblue!60!white, fill opacity=0.14] (A) -- (FTL) -- (FTR) -- cycle;
    \draw[fovblue!60!black, dashed, opacity=0.35]
        (A) -- (FTL) (A) -- (FTR) (A) -- (FBL) (A) -- (FBR);
    \draw[fovblue!60!black, dashed, opacity=0.35] (FTL) -- (FTR) -- (FBR) -- (FBL) -- cycle;
    \node[rotate=-34, fovblue!40!black, opacity=0.7, yshift=7pt] at ($(FTL)!0.5!(FTR)$) {\LARGE R \& FoV};

    \draw[->, blue, thick] (0,0,0) -- (0,0,4) node[right, black] {\Large $+z$};
    \draw[->, green!60!black, thick] (0,0,0) -- (0,9,0) node[right, black] {\Large $y$};
    \draw[->, red!80!black, thick] (0,0,0) -- (3,0,0) node[right, black] {\Large $+x$};
    \draw[->, red!80!black, thick] (0,0,0) -- (-2,0,0) node[above, black] {\Large $-x$};
    \draw[->, blue, thick] (0,0,0) -- (0,0,\shadowZ) node[below left, black] {\Large $-z$};


    \begin{scope}[shift={(0,0,0)}, x={(1cm,0cm)}, y={(0cm,1cm)}]
        \shade[ball color=egogreen, opacity=0.35] (0,0) circle (0.95cm);
    \end{scope}
    \node[egogreen] at (0, 1.0, 0.95) {\Large $R_{sum}$};
    \node at (-1.5, -1, 0) {\Large $P_e, V_e$};
    \pic at (0,0,0) {drone};

    \coordinate (Intruder) at (0, 4.5, 0);
    \begin{scope}[shift={(Intruder)}, x={(1cm,0cm)}, y={(0cm,1cm)}]
        \shade[ball color=obscrit, opacity=0.22] (0,0) circle (1.45cm);
        \shade[ball color=egogreen, opacity=0.25] (0,0) circle (0.95cm);
    \end{scope}

    \node[obscrit!80!black] at (0, 3.15, 1.5) {\Large $d_{crit}$};
    \begin{scope}[opacity=0.5]
        \pic at (Intruder) {drone};
    \end{scope}

    \coordinate (StarDetect) at (0, 6.5, 3.5);
    \coordinate (StarTrack) at (0, 5.2, 2.0);

    \node[circle, shading=ball, ball color=trackred, minimum size=0.56cm,
          inner sep=0pt] (rs1) at (StarDetect) {};
    \begin{scope}[shift={(StarDetect)}, x={(1cm,0cm)}, y={(0cm,1cm)}]
        \draw[trackred, thick] (-0.38,-0.38) rectangle (0.38,0.38);
    \end{scope}
    \node[trackred, above=0.14cm of rs1] {\large detection};
    \node[trackred, below right=0.14cm and 0.04cm of rs1] {\Large $P_i, V_i$};

    \begin{scope}[shift={(Intruder)}, x={(1cm,0cm)}, y={(0cm,1cm)}]
        \node[circle, shading=ball, ball color=trackred, minimum size=0.56cm,
              inner sep=0pt, opacity=0.45] (obs2) at (60:1.73cm) {};
        \draw[trajblue, thick] (60:1.45cm) -- (0,0);
        \fill[trajblue] (60:1.45cm) circle (0.06cm);
        \fill[trajblue] (0,0) circle (0.06cm);
        \node[trajblue, below right=0.02cm and 0.04cm of obs2] {\large trigger distance($d_t$)};
    \end{scope}
    \draw[-stealth, trackred, thick] (rs1) -- (obs2)
        node[midway, sloped, above] {\large tracking};

    \coordinate (TrajEnd) at (Intruder);
    \coordinate (AvoidStart) at (Intruder);
    \coordinate (AvoidEnd) at (0, 6.5, -1.0);

    \draw[-stealth, line width=1.2pt, dashed] (0,0,0) -- (TrajEnd);
    \node[rotate=9] at (0, 2, -0.3) {\large trajectory};

    \fill[trajblue] (AvoidStart) circle (0.07cm);
    \fill[black] (AvoidEnd) circle (0.07cm);
    \draw[-stealth, line width=1.2pt, dashed, trajblue]
         (AvoidStart) to[out=-50, in=200] (AvoidEnd);
    \node[trajblue] at (0, 5.5, -1.6) {\large avoidance};

    \draw[-stealth, line width=1.2pt, dashed] (AvoidEnd) -- (0, 8.7, 0)
        node[midway, sloped, below] {\large trajectory};

    \def\TLZ{-4.5}
    \def\TLZEnd{-3.8}

    \draw[densely dotted] (0,0,\shadowZ) -- (0, 0, -5.6);
    \draw[densely dotted] (0, 1.9, \shadowZ) -- (0, 1.9, -5);
    \draw[densely dotted] (0, 3.1, \shadowZ) -- (0, 3.1, -5);
    \draw[densely dotted] (0, 4.5, 0) -- (0, 4.5, -5.6);
    \draw[densely dotted] (AvoidEnd) -- (0, 6.5, -5);

    \draw[|-|, dashed, line width=1.5pt] (0, 0, \TLZEnd) -- (0, 3.1, \TLZEnd);
    \node[above] at (0, 1.55, \TLZEnd) {\Large $t_{end}$};

    \draw[|->, line width=1.5pt] (0, 0, \TLZ) -- (0, 1.9, \TLZ) node[midway, below] {\Large $t_{det}$};
    \draw[|->, line width=1.5pt] (0, 1.9, \TLZ) -- (0, 3.1, \TLZ) node[midway, below] {\Large $t_{avoid}$};
    \draw[|->, line width=1.5pt] (0, 3.1, \TLZ) -- (0, 4.5, \TLZ) node[midway, below] {\Large $t_{dec}$};
    \draw[|->, line width=1.5pt] (0, 4.5, \TLZ) -- (0, 6.5, \TLZ) node[midway, below] {\Large $t_{evade}$};

    \draw[|-|, dashed, line width=1.5pt] (0, 0, -5.6) -- (0, 4.5, -5.6);
    \node[below, yshift=-6pt] at (0, 2.25, -5.6) {\Large $t_{lat} = t_{end} + t_{dec}$};


    \begin{scope}[shift={(0, 9.5, 4)}]
        \node[circle, shading=ball, ball color=trackred, minimum size=0.62cm,
              inner sep=0pt] at (0,0,0) {};
        \node[trackred, align=center, right=0.2cm] at (0,0,0) {\large obstacle\\[-0.5ex]\large outside\\[-0.5ex]\large range};
    \end{scope}

\end{tikzpicture}}
  \caption{Sketch of dynamic obstacle avoidance of a UAV flying along a trajectory with nominal control input. The green circle around the UAV represents the aggregated spatial uncertainty of the UAV and the obstacle ($R_{sum}$), which is inflated to the critical distance $d_{crit}$ (orange circle) to account for perception and actuation time limitations for safe avoidance. Hence, safe collision avoidance requires triggering the evasive action beyond the trigger distance $d_t > d_{crit}$ such that sufficient actuation time $t_{evade}$ for evading the obstacle is available. The sensor's field of view (FoV) and range $R$ are shown in blue. The timeline shows the processing latency per RADAR frame $t_{end}=t_{det}+t_{avoid}$, the decision latency $t_{dec}$ from the first track to the trigger, the latency to the avoidance start $t_{lat}=t_{end}+t_{dec}$, and the time $t_{evade}$ to evade.
  }
  \Description{A sketch illustrating a time model for the assessment of the DOA in UAVs.}
  \label{fig:sys}
\end{figure}

Recently, millimeter-wave (mmWave) frequency modulated continuous wave (FMCW) radio detection and ranging (RADAR) is gaining popularity in UAV perception~\cite{mandaokar_widroit2025,hugler2018radar,corradi2022radar, soumya2023recent}. RADAR is computationally inexpensive and cost-effective compared to a LIDAR and stereo camera setup. It provides direct measures of range, velocity, and position~\cite{pointcloudoa2022}. However, it provides a sparse pointcloud, which makes accurate object size estimation difficult and results in larger position errors~\cite{pursueradar2020}. For aerial platforms, the mmWave RADAR has been used to perform accurate landing~\cite{radar_event_land2025}, to estimate the propeller speed~\cite{event_propeller2026}, and to assist soil sensing~\cite{radar_soil2026}. RADAR has also been studied for navigation, tracking, and detection, demonstrating its feasibility on medium-sized UAV platforms~\cite{batmobility2023,pursueradar2020, fusionradar2024,zhao2019mid}. However, the use of RADAR for fast, computationally efficient onboard DOA at more than 50\,Hz has not been studied or validated.

This work analyzes both latency and spatial bounds for successful DOA (see Figure~\ref{fig:sys}) and reports on our system-level implementation of a mmWave FMCW 77-81 GHz RADAR-based DOA onboard a resource-limited UAV. The DOA timing can be partitioned into detection latency $t_{det}$, avoidance latency $t_{avoid}$, decision latency $t_{dec}$, and evasion latency $t_{evade}$, where $t_{det}$ is the aggregate execution time of the obstacle detection and tracking modules.
$t_{avoid}$ is the aggregate execution time of the avoidance modules. $t_{dec}$ defines the time to trigger the avoidance starting from the first track of the obstacle given by the tracker. Finally, $t_{evade}$ represents the time required for the UAV to overcome its inertia and achieve the lateral displacement necessary for collision avoidance under its acceleration budget.
The critical distance $d_{crit}$ defines the boundary for successful DOA. If avoidance is triggered before the obstacle enters $d_{crit}$, the UAV can still evade.  $d_{crit}$ is calculated based on $t_{end}$ and the closing speed $v_c$ of the obstacle wrt.~the UAV.

This research builds upon distributionally robust acceleration control barrier filter (DR-ACBF)~\cite{dracbf2026}, combining efficient RADAR-based detection~\cite{mandaokar_widroit2025} and  tracking~\cite{mandaokar_safeimm2025}. Based on the integration of these three components, this work provides the following novel contributions: 

\begin{enumerate}
    \item An analysis of temporal and spatial constraints of RADAR perception for fast DOA with the decision latency, moving obstacles, a continuous-clearance condition, and a trigger design rule.

    \item A mmWave RADAR 100\,Hz pointcloud stream implementation for low-RCS non-metallic obstacles integrated with motion-compensated tracking that combines the tracks with the full attitude and velocity of the UAV for fast forward flights.

    \item An efficient system implementation and validation of DOA on a workstation without a GPU and a Raspberry Pi 4B (Pi4B) onboard companion computer, with indoor and outdoor flights and ablations of the RADAR rate, the tracker covariance, and the platform acceleration.

    \item A data set with different object sizes under different visibility conditions.
    
\end{enumerate}

The system-level insight from our experiments is that sensing latency is not the limiting factor for fast DoA. On a Raspberry Pi4B, RADAR processing takes 13.65\,ms, which is minor relative to the decision latency (median 61\,ms) and the time required to execute an evasive maneuver (0.33\,s). Therefore, the safe operating speed is governed mainly by detection range, accurate velocity estimation, and UAV agility, but not alone by sensing latency.

\section{Related Work}

State-of-the-art approaches address fast DOA by providing low-latency perception, while accounting for UAV acceleration limits. These approaches rely on event cameras, RGB-D cameras, or small LIDARs and perform well under suitable conditions, but require high computational power or are expensive. Following the setting in Figure~\ref{fig:sys}, DOA challenges can be summarized into detection latency, detection range, decision latency, avoid latency, and evade latency. The detection range $R$ specifies the maximum distance at which obstacles can be robustly detected. A small $R$ leaves less time to trigger avoidance, and slow sampling and detection increase $t_{det}$. Trackers that need many frames for a reliable velocity estimate increase $t_{dec}$, and complex control algorithms increase $t_{avoid}$. UAVs with large inertia and limited actuation prolong $t_{evade}$. Each of these delays the reaction.

To overcome \textit{detection latency} $t_{det}$ that slows agile flights, Falanga et al.~\cite{falangaral2019} introduced event-based perception~\cite{doafalanga2020} and fast trajectory planning~\cite{doafast2019}. However, visual degradation in darkness drastically reduces the reliability of vision sensors. LIDAR systems elude lighting issues and provide localization, which can improve DOA~\cite{FAPP2025} but add significant perception latency, while still struggling with dense smoke~\cite{pointcloudoa2022}. Similarly, RADAR was adapted for obstacle avoidance on UAVs in~\cite{radarMAV2021}, highlighting its superiority over optical sensors in dusty surroundings. BatMobility~\cite{batmobility2023} showcased mmWave RADAR-only navigation, focusing on static environments rather than high-speed dynamic objects. However, detecting small objects with low RADAR cross sections (RCS) and low reflectivity requires a tuned configuration and an efficient detection system~\cite{mandaokar_widroit2025}.

The detection latency also depends on the computational complexity of the detection model. Most camera-based detection models rely on neural networks (NNs), which require powerful GPUs for training and inference~\cite {rgbdsmalldoa2022, onboardsensinguav2021, rgbdoa2024}. Similar to a camera, LIDAR requires even more computing power to provide state and velocity measurements for the detection model, which is typically based on NNs~\cite{FAPP2025,pointcloudoa2022}. While LIDAR and camera outputs provide more accurate estimates of size (bounding box), class, and state, RADAR imposes a much lower computational burden on the detection model due to its inherent ability to measure velocity. However, many RADAR-based approaches still rely on NN-based obstacle size estimation~\cite {batmobility2023,fusionradar2024,kosuge2022, gupta2021}. Recent RADAR implementations~\cite{mandaokar_widroit2025} have demonstrated lightweight detection models with position accuracy under $0.5$\,m, sufficient for the DOA use case, along with effective avoidance strategies that account for measurement uncertainties arising from large position errors (e.g., DR-ACBF~\cite{dracbf2026}).

The \textit{detection range} $R$ of the studied camera perception~\cite{rgbdoa2024, rgbdsmalldoa2022,doafalanga2020} is limited, leaving even less time to react to a fast-moving obstacle. An event camera offers very low detection latency, as demonstrated by the DOA event camera in~\cite{doafalanga2020}. However, like many camera-based systems, the event camera needs multiple frames to estimate object speed, which increases the \textit{decision latency} $t_{dec}$ and offsets its low latency~\cite{rgbdoa2024, rgbdsmalldoa2022}. While LIDAR offers long range and high resolution, its NN-based detection and multi-frame velocity estimation requirements limit its use for DOA in small- to medium-sized UAVs~\cite{FAPP2025,pointcloudoa2022}. RADAR offers the longest detection range, a lightweight body, and direct velocity measurements, but suffers from poor detection resolution and high noise in cluttered environments~\cite{mandaokar_widroit2025}.

Fast DOA requires a small \textit{avoidance latency} $t_{avoid}$ to further reduce $t_{end}$. $t_{avoid}$ depends on the control method, as control variable optimization and command generation add computation time. Existing approaches use model predictive controllers (MPCs), which are slower but more accurate than faster reactive methods~\cite{doafalanga2020}. Reactive methods, such as artificial potential fields, work well for static or slowly moving scenes. Recently, control barrier functions (CBFs) have gained popularity for providing safety in highly dynamic environments. CBFs enforce rigid forward invariance of safe sets~\cite{ames2017tac2017,HOCBF12024}.  The DR-ACBF provides low-latency reaction and accounts for measurement uncertainties to perform a fast DOA, while accounting for perception and actuation delays~\cite{dracbf2026}. 

Finally, the \textit{evade latency} $t_{evade}$ measures the UAV platform's limit on how quickly it can react to the provided avoidance commands. It depends on the UAV's size, weight, and maximum allowable acceleration~\cite{falangaral2019}.
Falanga et al.~\cite{falangaral2019} investigated UAV actuation limits for the DOA with respect to perception latency from event, mono, and stereo frames, considering the detection range and the maximum allowable acceleration required for safe avoidance. Researchers have studied different onboard cameras and LIDAR to perform efficient DOA, such as \cite{falangaral2019,doafalanga2020,rapidoaicra2025} for an event camera,~\cite{rgbdsmalldoa2022,rgbdoa2024} for RGBD camera, and~\cite{onboardsensinguav2021,FAPP2025,flydyn2025} LIDAR, 
but no prior work presents RADAR-based onboard perception to perform DOA in UAVs. Onboard RADAR perception is currently limited to detection~\cite{fang2025cubedn}, slow obstacle avoidance~\cite{radarUAV2019,radarMAV2021,batmobility2023,saouav2024}, drone pursuing~\cite{pursueradar2020} or fusion~\cite{fusionradar2024,yu2020autonomous, huang2021improved, deng2022}, which is hindered by the slow sampling rate, heavy detection models, and poor control strategies.

\section{System Model}
\label{sec:system}

This section presents our system model and assumptions for DOA as well as the computation of the key latencies $t_{det}$, $t_{avoid}$, $t_{dec}$, and $t_{evade}$. We assess the perception limits for DOA with the estimated $t_{end}$ at the preferred RADAR sampling rate. $d_{crit}$ is then estimated based on $t_{end}$ and the assumed closing speed $v_c$. These estimates also provide reference bounds for evaluating the experimental results. Falanga et al.~\cite{falangaral2019} relate the maximum speed of a UAV to its perception latency for a static obstacle and a single sensing instant. We extend this analysis in four ways that matter for fast DOA with RADAR: (i) the latency includes the decision latency of the tracker and the trigger, (ii) both the UAV and the obstacle move, which~\cite{falangaral2019} discuss but do not analyze, (iii) the safety condition holds continuously over the encounter instead of only at the time of crossing, and (iv) the trigger threshold is linked to the critical distance.

\subsection{Latency}

Every RADAR frame is processed within $t_{end} = t_{det}+t_{avoid}$. The avoidance command, however, is issued only in the frame in which the trigger fires. If it fires at the first track of the obstacle, the command follows after $t_{end}$; otherwise, the trigger evaluates the following frames while the tracker refines the obstacle state, and the command follows after the additional decision latency $t_{dec}$ from the first track to the trigger. The latency from first detection to the avoidance command is therefore

\begin{equation}
t_{lat} = t_{end} + t_{dec}.
\label{eq:tlat}
\end{equation}

Our model differs from~\cite{falangaral2019} in that we explicitly account for the decision latency $t_{dec}$ rather than assuming the maneuver is initiated immediately upon detection, i.e., $t_{dec} = 0$. In our system, the trigger may fire immediately or after several frames, so $t_{dec} \geq 0$. RADAR-based velocity estimation keeps $t_{dec}$ small because it uses the obstacle's velocity measurements from the first track, unlike camera-based sensing.

\subsection{Allowable Limit}

Following~\cite{falangaral2019}, consider a UAV flying at speed $v$ toward a static obstacle detected at range $R$; it reaches the obstacle after $R/v$. Within this time, the UAV must react ($t_{lat}$) and move sideways by $R_{sum}$. With the lateral acceleration $a$, accelerating and then braking to a stop takes $2\sqrt{R_{sum}/a}$, so the allowable UAV speed is

\begin{equation}
v_{\max}(R, a, t_{lat}) = \frac{R}{t_{lat} + 2\sqrt{R_{sum}/a}}.
\label{eq:vmax_mmwave}
\end{equation}

Eq.~\eqref{eq:vmax_mmwave} is adopted from~\cite{falangaral2019} and used to derive limits of the maximum speed and the critical distance in Table~\ref{tab:exp_latency}, which is further used to compare with the camera latencies reported in~\cite{falangaral2019}. 

\subsection{Critical Distance and Continuous Clearance}

Consider a UAV with velocity $V_e$ and an obstacle with velocity $V_i$, closing at $v_c$, i.e., the magnitude of $V_i - V_e$ along the line of sight (LoS). Let $b$ be the miss distance of their relative trajectory, i.e., the closest distance at which the obstacle would pass without avoidance ($b = 0$ for a head-on encounter). All distances are center-to-center, and a collision occurs if the distance falls below $R_{sum}$. Avoidance requires the lateral displacement $y_{req} = \max(0, R_{sum} - b)$, which the UAV achieves with the acceleration budget $a$ after

\begin{equation}
t_{evade} = \sqrt{2\,y_{req}/a},
\label{eq:tevade}
\end{equation}

\noindent i.e., the UAV accelerates laterally at $a$ until $y_{req}$ is reached. The critical distance is

\begin{equation}
d_{crit} = R_{sum} + v_c \left(t_{lat} + t_{evade} \right).
\label{eq:critical_distance}
\end{equation}

If the avoidance is triggered at $d_t > d_{crit}$, the UAV completes the lateral displacement $y_{req}$ before the obstacle has closed in to $R_{sum}$. If the UAV then maintains this displacement until the obstacle has passed, the distance remains above $R_{sum}$ throughout the encounter. This is why Eq.~\eqref{eq:critical_distance} contains the term $R_{sum}$; without it, the displacement would be completed only when the obstacle arrives, and the distance would drop below $R_{sum}$ before that. Our controller holds the displacement for 1.2\,s. The head-on case ($b = 0$) requires the largest displacement and is therefore the most conservative case, which we use in the evaluation. Only when ground truth is available, we use the measured $b$ for relaxing the conservative displacement.

\subsection{Trigger Design}

The system uses DR-conditional value-at-risk (DR-CVaR)~\cite{dracbf2026}, which fires when the predicted distance falls below the safety distance $D_{safe} = R_{sum} + v_c\,\tau$, where $\tau$ is a latency budget. The trigger fires before the obstacle enters $d_{crit}$ if $D_{safe} \ge d_{crit}$. Comparing with Eq.~\eqref{eq:critical_distance}, this holds if

\begin{equation}
\tau \ge t_{lat} + t_{evade},
\label{eq:tau}
\end{equation}

\noindent which turns the critical distance into a design rule for the trigger. In our experiments, we use $\tau = 1.5$\,s. The trigger then fires 1.5\,s before the predicted collision, which is much longer than the time needed to react and evade. Additionally, the trigger requires a sufficient sensor detection range.

\subsection{Ego Speed Limit}

Eq.~\eqref{eq:vmax_mmwave} assumes a static obstacle. For a moving obstacle, the closing speed also contains the obstacle speed, which reduces the admissible UAV speed. Requiring that an obstacle detected at the detection range $R$ can still be avoided in a head-on encounter, i.e., $d_{crit} \le R$, yields the maximum UAV speed

\begin{equation}
\|V_e\| \le \frac{R - R_{sum}}{t_{lat} + t_{evade}} - \|V_i\|.
\label{eq:vego}
\end{equation}

Consequently, every meter of detection range gains $1/(t_{lat} + t_{evade})$ of admissible closing speed. These parameters enable an analysis of the limits for safe DOA. We assume the following conditions for this analysis: a sensor sampling rate of 100\,Hz, resulting in $t_{det}=10\,$ms; a control loop at 200\,Hz, resulting in $t_{avoid} = 5\,$ms (i.e., $t_{end}=$15\,ms); and closing velocity $v_{c} = v_{max}$. We are particularly interested in short-range perception and adopt the values for the range $R$ and the lateral acceleration budget $a$ from~\cite{falangaral2019}. Table~\ref{tab:exp_latency} shows the computed limits for $v_{max}$ and the $d_{crit}$ for a minimum safety radius $R_{sum} = 0.8\,$m, with $v_c = v_{\max}$ (static obstacle), $t_{lat} = t_{end}$ (single-frame detection and immediate trigger), and $t_{evade}$ from Eq.~\eqref{eq:tevade} with $b = 0$.

\begin{table}[H]
\centering
\caption{Limits on $v_{max}$ (Eq.~\eqref{eq:vmax_mmwave}) and $d_{crit}$ (Eq.~\eqref{eq:critical_distance}) for DOA considering different perception ranges and acceleration budgets.}
\label{tab:exp_latency}
\resizebox{\columnwidth}{!}{%
\begin{tabular}{cccccccccccc}
\toprule
\textbf{Range $R$} & \textbf{$t_{\text{end}}$} & \multicolumn{2}{c}{$a{=}10$} & \multicolumn{2}{c}{$a{=}25$} & \multicolumn{2}{c}{$a{=}50$} & \multicolumn{2}{c}{$a{=}200$} \\
\cmidrule(lr){3-4} \cmidrule(lr){5-6} \cmidrule(lr){7-8} \cmidrule(lr){9-10}
(m) & (s) & $v_{\max}$ & $d_{crit}$ & $v_{\max}$ & $d_{crit}$ & $v_{\max}$ & $d_{crit}$ & $v_{\max}$ & $d_{crit}$  \\
\midrule

2 & 0.015 & 3.44 & 2.23 & 5.37 & 2.24 & 7.46 & 2.25 & 14.14 & 2.28 \\
4 & 0.015 & 6.89 & 3.66 & 10.73 & 3.68 & 14.93 & 3.69 & 28.27 & 3.75 \\
5 & 0.015 & 8.61 & 4.37 & 13.41 & 4.39 & 18.66 & 4.42 & 35.34 & 4.49 \\
8 & 0.015 & 13.78 & 6.52 & 21.46 & 6.55 & 29.85 & 6.59 & 56.54 & 6.71 \\
\bottomrule
\end{tabular}
}
\end{table}

Under these identical conditions, the sensing modalities of~\cite{falangaral2019} can be compared directly: for $R = 2$\,m and $a = 10$\,m/s$^2$, Eq.~\eqref{eq:vmax_mmwave} yields 3.38\,m/s with their latency of a monocular camera (26\,ms), 3.43\,m/s with that of a stereo camera (17\,ms), and 3.49\,m/s with that of an event camera (7\,ms), while $t_{end} = 15$\,ms yields 3.44\,m/s. Since their latencies reflect only sensor timing, whereas $t_{end}$ includes processing, this comparison is conservative for RADAR. At these ranges, the geometric term $2\sqrt{R_{sum}/a}$ dominates latency, so the main benefit of RADAR is maintaining the sensing range $R$ in poor visibility rather than achieving lower latency.

Latency alone, however, does not ensure safety; the avoidance must also be triggered beyond $d_{crit}$. Table~\ref{tab:exp_latency} shows that $d_{crit}$ grows almost linearly with $R$, from 2.2--2.3\,m at $R = 2$\,m to 6.5--6.7\,m at $R = 8$\,m, and hardly depends on $a$. Eq.~\eqref{eq:vmax_mmwave} completes the displacement only at the moment of crossing, whereas $d_{crit}$ keeps the distance above $R_{sum}$. Therefore, an obstacle detected at $R = 2$\,m is already inside $d_{crit}$ at $v_{\max}$. 
From $R = 4$\,m, $R > d_{crit}$ holds with a margin of only 0.25--1.48\,m, and during any additional delay, such as $t_{dec}$, the obstacle comes $v_c\,t_{dec}$ closer and reduces this margin. For DOA, this implies that the state is evaluated using the inequality $d_t > d_{crit}$, where $d_t$ is the actual distance at which the DR-ACBF avoidance trigger occurs. If $d_t > d_{crit}$, clearance is ensured under the assumptions of Eq.~\eqref{eq:critical_distance} (constant closing speed, acceleration $a$ available at the command). Conversely, $d_t \leq d_{crit}$ does not imply a collision, but clearance is no longer ensured by Eq.~\eqref{eq:critical_distance}.

\section{Implementation}
\label{sec:implementation}

\begin{figure}
  \centering
  \includegraphics[width=\linewidth]{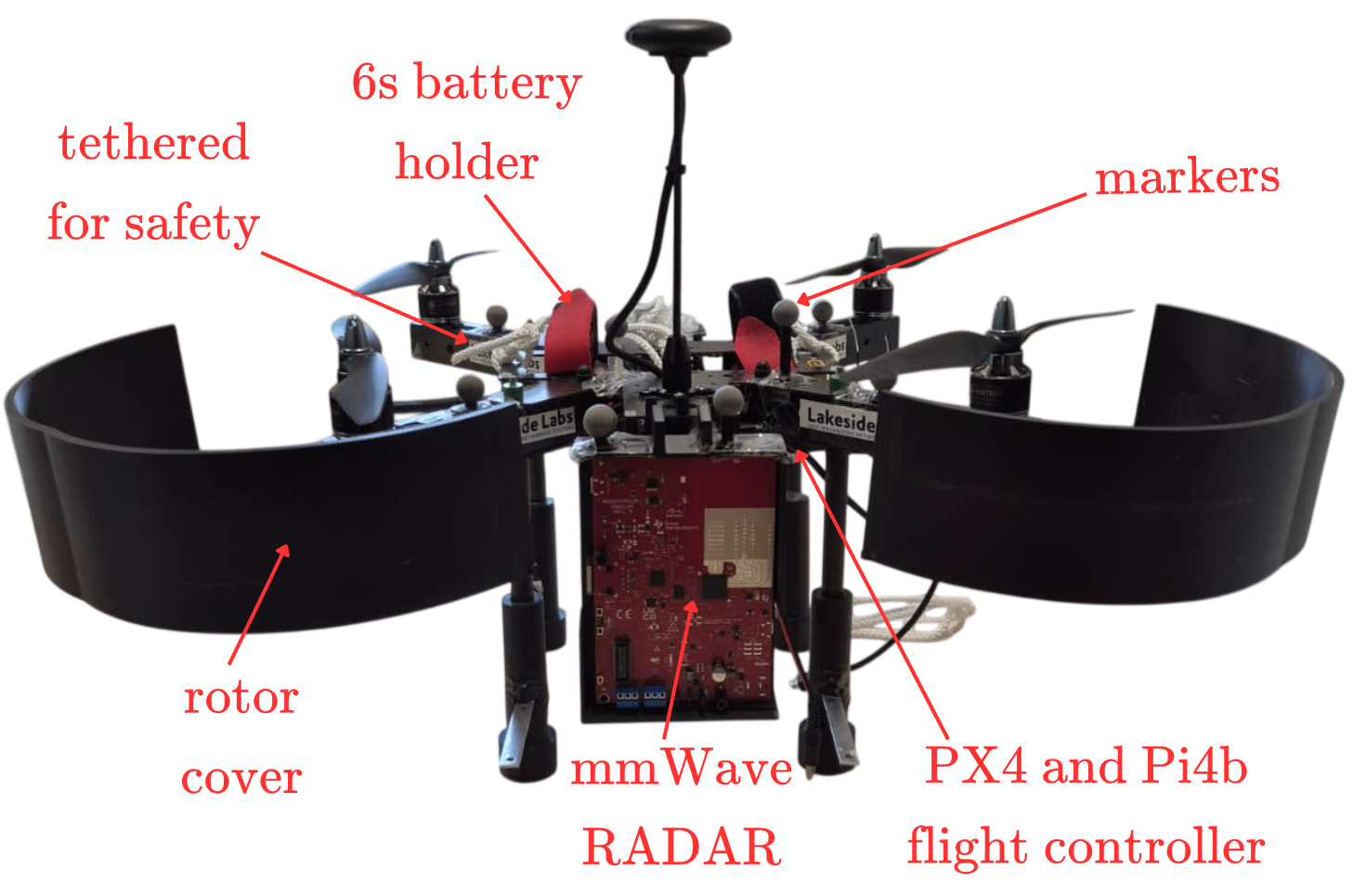}
  \caption{Experimental setup of our quadrotor with the mmWave RADAR sensor.}
  \Description{A quadrotor with mmWave TI AWR2944EVM RADAR board and Arducam global shutter camera sensor.}
  \label{fig:setup}
\end{figure}

This section presents the hardware and software stack used to validate the proposed mmWave RADAR-based DOA system.  Figure~\ref{fig:setup} shows the quadrotor platform with a Texas Instruments (TI) mmWave FMCW 77-81 GHz automotive-grade RADAR\footnote{\url{https://www.ti.com/tool/AWR2944PEVM}}, and Figure~\ref{fig:system} depicts the proposed system control loop. The pointcloud at 100\,Hz is fed into the detection model~\cite{mandaokar_widroit2025}, which uses a sliding window and density-based spatial clustering of applications with noise (DBSCAN)~\cite{ester1996density} to provide dynamic boxes of detected obstacles with $(x,y,z,vx,vy,vz)$ states. These states serve as input measurements for the SAFE-interacting multiple model (IMM) tracker~\cite{mandaokar_safeimm2025}, which is preferred for its lightweight, robust tracking. Our implementation runs the RADAR data stream and algorithmic processing in parallel. The pointcloud is received at 100\,Hz, and the detection and tracking always process the newest frame, so that the track rate depends on the processing time per frame.
The UAV receives obstacle tracks to enable avoidance using the DR-ACBF model~\cite{dracbf2026}, which incorporates the DR-CVaR trigger to initiate ACBF-based avoidance acceleration.

\begin{figure}
  \centering
  \resizebox{\linewidth}{!}{\begin{tikzpicture}[
    >=stealth,
    box/.style={draw, line width=1.2pt, rounded corners=10pt, align=center,
                minimum width=3.1cm, minimum height=1.5cm, font=\Large},
    flow/.style={line width=3pt, rounded corners=12pt,
                 -{Triangle[length=4mm, width=5.5mm]}},
    rate/.style={font=\Large}
]

    \node[box, minimum height=2.5cm] (radar) at (0,6.3) {};
    \begin{scope}[shift={(radar.center)}]
        \filldraw[black, rounded corners=2pt] (-0.45,0.45) rectangle (0.45,1.0);
        \draw[line width=2.2pt]          (0,0.42) ++(235:0.32) arc[start angle=235, end angle=305, radius=0.32];
        \draw[line width=2.2pt, gray!70] (0,0.42) ++(235:0.60) arc[start angle=235, end angle=305, radius=0.60];
        \draw[line width=2.2pt] (-0.62,-0.42) -- (0.62,-0.42);
        \node[font=\Large] at (0,-0.85) {RADAR};
    \end{scope}

    \node[box, minimum height=1.9cm] (det) at (0,2.9) {detection\\\&\\tracking};

    \node[box, minimum height=2.5cm] (mocap) at (0,-1.6) {};
    \begin{scope}[shift={($(mocap.center)+(0,0.1)$)}, line width=1.4pt]
        \draw (0,0.95) circle (0.13);                                  
        \draw (0,0.82) -- (0,0.18);                                    
        \draw (0,0.70) -- (-0.32,0.55)   (0,0.70) -- (0.32,0.55);      
        \draw (-0.32,0.55) -- (-0.44,0.20)  (0.32,0.55) -- (0.44,0.20);
        \draw (0,0.18) -- (-0.22,-0.18)  (0,0.18) -- (0.22,-0.18);     
        \draw (-0.22,-0.18) -- (-0.28,-0.55) (0.22,-0.18) -- (0.28,-0.55); 
        \foreach \p in {(0,0.70),(-0.32,0.55),(0.32,0.55),(-0.44,0.20),(0.44,0.20),
                        (0,0.18),(-0.22,-0.18),(0.22,-0.18),(-0.28,-0.55),(0.28,-0.55)}{
            \filldraw[fill=white] \p circle (0.062);
        }
        \node[font=\Large] at (0,-0.95) {mocap};
    \end{scope}

    \node[box, minimum width=2.9cm, minimum height=2.5cm] (uav) at (5.8,2.6) {};
    \begin{scope}[shift={($(uav.center)+(0,0.25)$)}, scale=1.15]
        \draw[line width=1.6pt, black!60, line cap=round] (0,0.02) -- (-0.36,0.16);
        \draw[line width=1.6pt, black!60, line cap=round] (0,0.02) -- ( 0.36,0.16);
        \draw[line width=1.6pt, black!60, line cap=round] (0,-0.02) -- (-0.42,-0.16);
        \draw[line width=1.6pt, black!60, line cap=round] (0,-0.02) -- ( 0.42,-0.16);
        \filldraw[gray!20, draw=black!55, line width=0.8pt]
            (-0.36,0.16) ellipse [x radius=0.19, y radius=0.065];
        \filldraw[gray!20, draw=black!55, line width=0.8pt]
            ( 0.36,0.16) ellipse [x radius=0.19, y radius=0.065];
        \filldraw[gray!20, draw=black!55, line width=0.8pt]
            (-0.42,-0.16) ellipse [x radius=0.22, y radius=0.075];
        \filldraw[gray!20, draw=black!55, line width=0.8pt]
            ( 0.42,-0.16) ellipse [x radius=0.22, y radius=0.075];
        \filldraw[black!80] (0,-0.02) ellipse [x radius=0.17, y radius=0.09];
        \filldraw[black!55] (0, 0.045) ellipse [x radius=0.13, y radius=0.065];
    \end{scope}
    \node[font=\Large] at ($(uav.center)+(0,-0.75)$) {uav};

    \node[box, minimum height=2.5cm] (fc) at (10.2,2.6) {};
    \node[fill=black, text=white, rounded corners=5pt, inner xsep=8pt, inner ysep=4pt,
          font=\Large\bfseries\itshape, align=center]
          at ($(fc.center)+(0,0.5)$) {PX4};
    \node[font=\Large, align=center] at ($(fc.center)+(0,-0.5)$) {flight\\controller};

    \node[diamond, draw, line width=1.2pt, left color=red!85!black, right color=orange,
          text=white, font=\Large, minimum width=3cm, minimum height=2.6cm,
          inner sep=1pt] (trig) at (14.6,2.6) {trigger};

    \node[box] (avctrl) at (14.6,6.3) {avoidance\\controller};
    \node[box] (avcmd)  at (10.2,6.3) {avoidance\\command};

    \node[box] (wp) at (10.2,-0.6) {way\\points};

    \node[box, minimum width=2.4cm] (cp) at (3.4,-1.6) {current\\position};

    \draw[flow] (radar.south) -- (det.north)
        node[rate, midway, right=2mm] {100Hz};
    \draw[flow] (det.east) -- (det.east -| uav.west)
        node[rate, pos=0.45, above=1mm] {100Hz};
    \draw[flow] ($(mocap.north)+(1.0,0)$) |- ($(uav.west)+(0,-0.95)$)
        node[rate, pos=0.72, below=1mm] {180Hz};
    \draw[flow] (uav.east) -- (fc.west);
    \draw[flow] (fc.east) -- (trig.west)
        node[rate, midway, above=1mm] {200Hz};
    \draw[flow, red!80!black!90!white, draw=red!75] (trig.north) -- (avctrl.south);
    \draw[flow, draw=red!75] (avctrl.west) -- (avcmd.east);
    \draw[flow, draw=red!75] (avcmd.west) -| (uav.north);
    \draw[flow, draw=green!70!black!60!white] (trig.south) |- (wp.east);
    \draw[flow, draw=green!70!black!60!white] (wp.west) -| ($(uav.south)+(0.55,0)$);
    \draw[flow, rounded corners=6pt] ($(uav.south)+(-0.25,0)$) |- (cp.east);
    \draw[flow] (cp.west) -- (mocap.east);

    \draw[dashed, line width=1.6pt, gray, rounded corners=16pt]
        (-2.1,-3.3) rectangle (17.3,8.1);
    \node[gray, font=\Large\bfseries] at (7.6,7.75) {control loop};

\end{tikzpicture}}
  \caption{UAV control loop of the proposed DOA system. The mmWave RADAR provides a pointcloud at 100\,Hz, and the detection and tracking subsystem provides the position and velocity of the obstacle from the newest frame. The UAV uses either MOCAP for indoor or GPS \& range finder for outdoor to continuously update its position. The UAV runs the control loop at 200\,Hz on the Raspberry Pi4B with a PX4 flight controller. In the case of avoidance (red arcs), the avoidance controller generates an avoidance command; otherwise (green arcs), the UAV approaches the next waypoints.}
  \Description{Illustration of control to fly and perform avoidance.}  
  \label{fig:system}
\end{figure}

We validate the system on two computer platforms: a hardware-in-the-loop (HiL) stationary experiment on a laptop and a flying experiment with a Raspberry Pi4B companion computer. The laptop has 16\,GB RAM, an Intel i7 processor, and a 512\,GB SSD, while the Pi4B has 4\,GB RAM, an ARM Cortex-A72 processor, and a 128\,GB SD card for storage. Both systems operated on Ubuntu 20.04 and used ROS2 Galactic. Our UAV system was powered by a 4S 6600\,mAh battery and utilized a Holybro Pixhawk 4 (PX4) autopilot. The UAV has a total weight, including the RADAR module, of 2.0\,kg, a size of $52 \times 52 \times 25$\,cm$^3$ with propellers attached, and a thrust-to-weight ratio of $\approx$3. The PX4 is configured for a maximum allowable acceleration $a=15$\,m/s$^2$.

\subsection{Sensor Configuration}
\label{subsec:RANGE}

The proposed RADAR profile is configured for frame settings of 3 TX ($N_{tx}$) and 4 RX ($N_{rx}$) for improved velocity resolution, carrier frequency $f_c=77$\,GHz, and 110 loops, giving 330 chirps per frame with 110 Doppler bins ($N_D$) per TX channel~\cite{TI_SWRA554A}. Constant false alarm rate (CFAR) is configured at 15\,dB threshold (range and Doppler branches), which limits clutter in the pointcloud output and provides fewer static points~\cite{rohling1983radar}. The configured velocity resolution $\Delta v$ and maximum radial velocity $v_{rad}$ is

\begin{equation}
\Delta v=\frac{c}{2f_cT_cN_DN_{tx}}\approx 0.295\,\text{m/s}
\label{eq:dv}
\end{equation}
\begin{equation}
v_{rad}=\frac{\Delta vN_D}{2}\approx\pm 16.2\,\text{m/s},
\label{eq:vrad}
\end{equation}

\noindent where $T_c$ is the chirp period (20~$\mu$s) and $T_f$ is the frame period (10~\,ms). This configuration enables the horizontal FoV to cover a wide azimuthal angle of roughly $\pm80^{\circ}$ and an elevation of $\pm20^{\circ}$, but we only consider detection in direct LoS of the UAV. For each chirp $S=57\,$MHz/$\mu$s with ADC sampling rate $f_s=11{,}130\,$ksps and $N=128$ ADC samples, the range resolution $\Delta R$  and configured maximum range $R_{max}$ is

\begin{equation}
\Delta R=\frac{c\,f_s}{2SN}\approx 0.23\,\text{m},
\qquad
R_{max}=\frac{c\,f_s}{2S}\approx 29.3\,\text{m}.
\end{equation}

This $R_{max}$ holds for complex ADC sampling, where all $N$ range bins are usable. Our configuration samples the ADC in real, which leaves only half of the $N$ range bins usable, so the theoretical limit is $\approx 14.6$\,m. Although the configured $R_{max}$ is higher, detectability heavily relies on the object's RCS~\cite{TI_RCS_mmWave}. For this study, we use three different ball sizes with diameters $\leq$0.3\,m, $\leq$0.2\,m, and $\leq$0.1\,m, with the smallest RCS of 0.001\,m$^2$. RADAR detection degrades for non-metallic objects, further reducing the configured detection range.

\subsection{Pointcloud Stream}
\label{subsec:stream}

A DOA system needs a fresh pointcloud with radial velocity in every frame. The onboard RADAR systems for UAV navigation and tracking in~\cite{radarMAV2021,batmobility2023,pursueradar2020} operate at lower pointcloud rates, and to the best of our knowledge, no prior onboard RADAR-based DOA system streams the 3D pointcloud at 100\,Hz. On the AWR2944~\cite{TI_AWR2944_userguide}, this rate is only reached without dropped frames if every stage of the pipeline, i.e., chirping, on-chip detection, transfer, and host processing, finishes within the frame period of 10\,ms. We optimized each stage for this:

\begin{figure}[H]
  \centering
  \resizebox{\linewidth}{!}{\begin{tikzpicture}[font=\footnotesize, >=Latex]
  \begin{scope}[x=0.3cm, y=0.45cm]
    \draw[->] (0,0) -- (27,0) node[right] {$t$ ($\mu$s)};
    \draw[->] (0,0) -- (0,4.6) node[above] {$f$};
    \fill[green!20] (7.5,0) rectangle (19.0,4.0);
    \draw[thick] (0,0.4) -- (5,0.4) -- (20,3.8) -- (20,0.4) -- (25,0.4) -- (26.5,0.74);
    \node[left] at (0,0.4) {77\,GHz};
    \node[right] at (20.3,2.6) {$S = 57$\,MHz/$\mu$s};
    \node at (13.25,3.65) {ADC: 128 samples};
    \foreach \x in {5,7.5,19,20} \draw[densely dotted] (\x,0) -- (\x,4.0);
    \draw[|<->|] (0,-0.55) -- node[below] {idle 5} (5,-0.55);
    \draw[|<->|] (5,-0.55) -- node[below] {2.5} (7.5,-0.55);
    \draw[|<->|] (7.5,-0.55) -- node[below] {$128/f_s = 11.5$} (19,-0.55);
    \draw[|<->|] (5,-1.65) -- node[below] {ramp end 15} (20,-1.65);
    \draw[|<->|] (0,-2.75) -- node[below] {$T_c = 20$} (20,-2.75);
    \node[above] at (13.5,4.7) {(a) one chirp};
  \end{scope}
  \begin{scope}[yshift=-4.4cm, x=0.78cm, y=0.5cm]
    \draw[->] (0,0) -- (10.8,0) node[right] {$t$ (ms)};
    \foreach \i in {0,...,10} {
      \pgfmathsetmacro{\k}{0.6*\i}
      \draw[fill=blue!15] (\k,0.6) rectangle ++(0.2,1.2);
      \draw[fill=blue!30] (\k+0.2,0.6) rectangle ++(0.2,1.2);
      \draw[fill=blue!50] (\k+0.4,0.6) rectangle ++(0.2,1.2);
    }
    \draw[fill=orange!25] (6.6,0.6) rectangle (10,1.8) node[midway, align=center] {on-chip\\detection};
    \draw[|<->|] (0,2.2) -- node[above] {3 TX $\times$ 110 chirps, 6.6\,ms} (6.6,2.2);
    \draw[|<->|] (6.6,2.2) -- node[above] {3.4\,ms} (10,2.2);
    \draw[densely dashed, gray] (10,0) -- (10,3.0);
    \foreach \x in {0,2,4,6,8,10} \node[below] at (\x,0) {\x};
    \draw[|<->|] (0,-1.1) -- node[below] {frame period $T_f = 10$\,ms (100\,Hz)} (10,-1.1);
    \node[above] at (5,3.6) {(b) one frame};
  \end{scope}
\end{tikzpicture}}
  \caption{RADAR configuration for 100\,Hz i.e., $t_f=$10\,ms. (a) One chirp: after the idle time and the ADC start time, the 128 samples at $f_s = 11.13$\,MHz end at 14.0\,$\mu$s of the 15\,$\mu$s ramp, so the chirp period of 20\,$\mu$s cannot be shortened much. (b) One frame: 3 TX $\times$ 110 chirps take 6.6\,ms and leave 3.4\,ms of each 10\,ms frame for the on-chip detection.}
  \Description{Illustration of RADAR frame configuration to achieve 100\,Hz sampling rate.}
  \label{fig:stream}
\end{figure}

\subsubsection{Chirping:} A frame consists of 3 TX $\times$ 110 chirps of $T_c = 20$\,$\mu$s (5\,$\mu$s idle and a 15\,$\mu$s ramp, Figure~\ref{fig:stream}a), i.e., 6.6\,ms of chirping, which leaves 3.4\,ms of each frame for the on-chip processing (Figure~\ref{fig:stream}b). The ramp is just long enough for the 128 ADC samples at 11.13\,MHz after the ADC start time of 2.5\,$\mu$s, and the short chirp keeps the velocity limit $v_{rad} = 16.2$\,m/s (Eq.~\eqref{eq:vrad}) above the expected speed of the obstacle. The 110 chirps per TX give $\Delta v = 0.295$\,m/s (Eq.~\eqref{eq:dv}).

\subsubsection{On-chip detection:} The CFAR runs with a 15\,dB threshold in range and Doppler, with peak grouping and without static clutter removal, and the sensor outputs only the detected points $(x, y, z, v_r)$ each with 16\,bytes, instead of range or Doppler maps.

\subsubsection{Transfer:} At 3.125\,Mbaud, the UART carries up to 3.1\,kB per 10\,ms frame. The point list is optimized to include a sufficient number of points across all RADAR streams, so that every frame is sent as the next is acquired and no backlog builds up.

\subsubsection{Host:} The driver polls the UART every 1\,ms, splits the byte stream into frames at the frame header, and publishes each frame as a pointcloud. Points with $|v_r| \ge 0.9\,v_{rad} \approx 14.6$\,m/s, close to the unambiguous limit, are treated as aliased and set to zero velocity; this also bounds the closing speed that the Doppler can report. The tracker subscribes with a queue depth of one, so that it always processes the newest frame and never works through a backlog of old frames.

\subsection{Detection}
\label{subsec:OD}

We employ a sliding-window algorithm combined with DBSCAN~\cite{mandaokar_widroit2025}. The detection model distinguishes static from dynamic points using radial-velocity thresholds and aggregates consecutive frames to output 3D bounding boxes representing candidate obstacles~\cite{mandaokar_widroit2025}. This velocity-based filtering significantly scales down the computational complexity of the subsequent DBSCAN clustering stage. $t_{det}$ includes the proposed detection model latency, which is reduced by a window size of 1 for candidate creation and dynamic filtering with a threshold of 0.5\,m/s. The DBSCAN clustering is set with $\epsilon$ of 0.5, and a minimum of 2 points, suited for the sparse RADAR pointcloud. We do not implement a camera or LIDAR baseline, since our goal is to show that RADAR alone enables DOA, including in dark and smoke where optical sensing degrades~\cite{radarMAV2021,pointcloudoa2022,mandaokar_widroit2025}.

\subsection{Tracking}
\label{subsec:OT}

We utilize a SAFE-IMM tracking architecture~\cite{mandaokar_safeimm2025}. The choice to utilize this architecture for RADAR perception is primarily driven by its ability to bound trajectory drift during model switches without incurring high computational costs. The SAFE-IMM architecture includes a post-fusion, covariance-aware gate that systematically evaluates the track's underlying Mahalanobis distance and uncertainty geometry, producing filtered state estimates $P_i$ and $V_i$, and the corresponding covariance $\Sigma_i$ for each obstacle. The SAFE-IMM tracking uses an efficient global nearest neighbor (GNN) to instantaneously associate tracks to multiple objects. In addition to the detection model latency, $t_{det}$ includes the tracker latency. Thus, to keep tracker latency low, the GNN uses a track-confirmation threshold of 1, a deletion threshold of 5, and SAFE-IMM with constant-velocity and constant-acceleration Kalman filters.

\subsection{Motion Compensation}
The RADAR is rigidly mounted on the UAV (see Figure~\ref{fig:setup}), so its measurements move and tilt with the quadrotor. In forward flight, the UAV pitches by tens of degrees, and static objects appear with a radial velocity of the order of the flight speed. We therefore combine the RADAR tracks with the attitude and velocity from the PX4 estimator at three levels before making them available for the avoidance trigger.

\subsubsection{Doppler compensation:}
Eqs.~\eqref{eq:dv}-\eqref{eq:vrad} concern the radial velocity $v_r$ of each point. A static point along the unit vector $\mathbf{u}$ in the RADAR frame appears with $v_r = -\mathbf{u}^\top \mathbf{v}_e^{r}$, where $\mathbf{v}_e^{r}$ is the UAV velocity in the RADAR frame. Before the dynamic-point filter, each point is compensated as

\begin{equation}
v_r^{c} = v_r + \mathbf{u}^\top \mathbf{R}_{rb}^\top \mathbf{R}(\mathbf{q})^\top V_e,
\label{eq:dopplercomp}
\end{equation}

\noindent where $V_e$ is the UAV velocity in the world frame, $\mathbf{R}(\mathbf{q})$ is the rotation from body to world given by the full attitude quaternion $\mathbf{q}$ (roll, pitch, and yaw), and $\mathbf{R}_{rb}$ is the fixed rotation from the RADAR to the body frame. Using only the yaw would leave a residual of order $\|V_e\|(1 - \cos\theta)$ at tilt $\theta$. The forward flights reach tilt angles of 35-64$^\circ$. Even at 5\,m/s and $\theta = 45^\circ$, the residual reaches 1.5\,m/s, causing static clutter to appear as false obstacles. After compensation, $v_r^{c}$ is the radial velocity of the point over the ground.

\subsubsection{Cartesian velocity measurement:} 
The RADAR measures only $v_r$, whereas the tracker and the avoidance operate on Cartesian velocities. We therefore map the compensated radial velocity $v_{r,j}^{c}$ of each point $j$ onto the LoS to the centroid $\bar{\mathbf{p}} = (\bar x, \bar y, \bar z)$ of the points of the frame, with azimuth $\alpha = \operatorname{atan2}(\bar y, \bar x)$ and elevation $\varepsilon$:

\begin{equation}
\mathbf{v}_j = v_{r,j}^{c}\,\left[\cos\alpha,\ \sin\alpha,\ \tan\varepsilon\right]^{\top}.
\label{eq:cartvel}
\end{equation}

Horizontally, $\mathbf{v}_j$ lies along the LoS; vertically, the factor $1/\cos\varepsilon$ in $\tan\varepsilon = \sin\varepsilon/\cos\varepsilon$ stays below 1.07 within the $\pm20^\circ$ elevation FoV. Points with $|v_{r,j}^c| < 0.5$\,m/s are discarded as static. The remaining points are clustered by DBSCAN, and each detection is the mean of its points in position and velocity. The SAFE-IMM uses the full measurement vector in its constant-velocity and constant-acceleration filters,

\begin{equation}
\mathbf{z} = [x,\ y,\ z,\ v_x,\ v_y,\ v_z]^\top,
\label{eq:meas}
\end{equation}

\noindent so the measured velocity enters the track from the first frame. The components that $v_r$ does not contain, i.e., the tangential motion and the vertical motion of a ball under gravity, are inferred by the tracker. Eqs.~\eqref{eq:dv}-\eqref{eq:vrad} bound the measured velocity and radial speeds; above the limit of Section~\ref{subsec:stream}, they are not measured.

\subsubsection{Ground-referenced tracks:}
The tracker estimates the position $\mathbf{p}_i^{r}$ relative to the RADAR and the velocity $\mathbf{v}_i^{r}$ over the ground. The avoidance transforms them into the world frame with the attitude at the time of the update,

\begin{equation}
P_i = P_e + \mathbf{R}(\mathbf{q})\mathbf{R}_{rb}\,\mathbf{p}_i^{r}, \qquad V_i = \mathbf{R}(\mathbf{q})\mathbf{R}_{rb}\,\mathbf{v}_i^{r},
\label{eq:track2world}
\end{equation}

\noindent so that the relative state $\hat P = P_i - P_e$, $\hat V = V_i - V_e$ contains the UAV motion exactly once. Since $\mathbf{p}_i^{r}$ is measured relative to the UAV, the GNSS position error does not enter $\hat P$. The closing speed is the projection of the relative velocity onto the LoS $\mathbf{n} = \hat P/\|\hat P\|$,

\begin{equation}
v_c = -\mathbf{n}^\top \hat V,
\label{eq:vc}
\end{equation}

\noindent which is used in the trigger and the DR-ACBF, and in Eq.~\eqref{eq:critical_distance}. It contains $v_z$, so the vertical motion enters the risk prediction and the trigger, whereas the avoidance acceleration is horizontal: its vertical component is set to zero, and the altitude is held by the position setpoint.
Eq.~\eqref{eq:cartvel} assumes motion along the LoS, which holds for approaching obstacles; for crossing obstacles, the velocity relies on the positions.

\subsubsection{Obstacle gating and timing:}
An obstacle enters the trigger only if it closes in, $v_c \ge v_{min}$ (Eq.~\eqref{eq:vc}), and moves over the ground, $\|V_i\| \ge v_{min}$, with $v_{min} = 1$\,m/s. The first gate removes receding objects; the second removes static clutter and slowly moving objects such as walking people, which can pass the first gate in forward flight because $v_c$ contains the UAV speed. The trigger is evaluated on every tracker output, and the first track output of an obstacle is used directly, while the 200\,Hz control loop sends the latest command to the PX4. During a dodge, the position along the flight path is held at the trigger point, so that in forward flight, the UAV also brakes while it moves sideways. The DR-ACBF is re-evaluated every 20\,ms with the current UAV state. The PX4 follows a smooth lateral reference, accelerating during the first half of the displacement and decelerating during the second. So that the UAV stops at the dodge position rather than overshooting. Afterward, it rejoins its nominal path. Together, these steps make the avoidance independent of flight speed up to the limit given by Eq.~\eqref{eq:vego}.

\subsection{Avoidance}
\label{subsec:avoidance}

Instead of solving a heavy quadratic program, the DR-ACBF leverages a Gauss-Southwell deterministic projection~\cite{dracbf2026,fastproj2023,gauss2025} to find the closest safe acceleration inside a linear half-space. The closed-form analytical boundary is dynamically adjusted using the DR-CVaR early-warning margins, which act as triggers (see Figure~\ref{fig:system}). Based on the estimated relative state of the obstacle, DR-CVaR triggers avoidance, and the DR-ACBF computes the evasive acceleration. $t_{avoid}$ aggregates the time taken by the DR-CVaR trigger, ACBF, and Gauss-Southwell projection to provide the evasive acceleration.

For each obstacle $i$, with relative position $\hat P_i = P_i - P_e$, $\mathbf{n}_i = \hat P_i/\|\hat P_i\|$, distance $d_i = \|\hat P_i\|$, and closing speed $v_{c,i} = \max(0, v_c)$ of obstacle $i$ (Eq.~\eqref{eq:vc}), the DR-ACBF requires the predicted distance after the horizon $H$ to stay above an effective radius, which is linear in the commanded acceleration $a_e$~\cite{dracbf2026}:

\begin{equation}
\mathbf{n}_i^\top a_e \le \frac{d_i - R_{\mathrm{eff},i} - v_{c,i} H - \lambda_i\,\sigma_i}{H^2/2},
\label{eq:acbf}
\end{equation}

with $R_{\mathrm{eff},i} = R_{sum} + d_{cl} + v_{e,max}\Delta + \tfrac{1}{2}a\Delta^2 + \tfrac{1}{6}j_{max}H^3 + \tfrac{1}{2}\hat a_i H^2$, where $\Delta$ accounts for the actuation delay and $\hat a_i$ is the obstacle acceleration envelope of the sliding mode differentiator (SMD). Following~\cite{dracbf2026}, measurement uncertainty enters through the Cantelli factor $\lambda_i = \sqrt{(1-\alpha_i)/\alpha_i}$, with $\alpha_i$ the share of $\alpha_{total}$ allocated to obstacle $i$, and $\sigma_i^2 = \mathbf{n}_i^\top(\Sigma_e + \Sigma_i)\mathbf{n}_i + (H\sigma_v)^2$, the position covariance of the tracker projected onto the LoS plus an optional closing-speed uncertainty $\sigma_v$ propagated over the horizon. The horizon is $H \ge \max(0.15\,\text{s}, t_{ca} - \Delta)$, where $t_{ca}$ is the predicted time of the worst risk from the trigger. The Gauss-Southwell projector finds the acceleration closest to the nominal command that satisfies Eq.~\eqref{eq:acbf} for all obstacles and the $\|a_e\|_\infty \le a$. If the constraints are not satisfied after the iterations, the last iterate is applied, i.e., a best-effort saturated command. Otherwise, a failsafe is issued to land safely.

\section{Results}
\label{sec:result}

This section presents the experimental validation of the proposed mmWave RADAR DOA framework, including detection accuracy, detection rate, trigger rate, DR-ACBF performance at different ranges, and resource efficiency. The experimental validation is performed in two settings: a stationary HiL experiment to assess detection capabilities, with 390 throws across three ball sizes, and an onboard experiment validation using a Pi4B as the companion computer with 20 tethered indoor flights and untethered outdoor flights (15 hovering avoidance, 15 in forward flight at 2, 3.5, and 5\,m/s, and 9 for the uncertainty ablation). For all the experiments, we set the time horizon $H=1.0$\,s, the radius of the UAV and obstacle as 0.5\,m and 0.3\,m respectively (i.e., $R_{sum} = 0.8\,$m), a maximum UAV acceleration $a = 15.0\,m/s^2$ and a maximum ego velocity $v_{e,max} = 7.0\,$m/s. The DR-ACBF relies on a total risk budget of $\alpha_{total} = 0.15$, with a CVaR risk sensitivity of $\alpha = 0.01$. Obstacle acceleration envelopes are estimated using an SMD differentiator with parameters $(\gamma, L_0, L_1, L_2) = (1.5, 4.0, 3.0,2.0)$ and a regularization coefficient of $\alpha_{smd} = 0.01$. For obstacle clearance, we set a lateral offset of $d_{cl} = 1.0\,$m and a Wasserstein radius of $\epsilon_{wass} = 0.05$ to ensure numerical robustness. The safety-critical constraints are satisfied via a fixed-time Gauss-Southwell projector, utilizing 12 iterations plus one per obstacle, a relaxation factor of $1.0$, and a projection tolerance $\epsilon_{proj} = 1 \times 10^{-9}$. The DR-CVaR trigger uses the safety distance $D_{safe}$ of Section~\ref{sec:system} with $\tau = 1.5$\,s, which satisfies Eq.~\eqref{eq:tau} with a large margin. Since $D_{safe}$ exceeds the detection range of the balls, the trigger fires without waiting for an approaching ball to come closer. $\Delta = 0.1$\,s is a delay cushion in the DR-ACBF (Section~\ref{subsec:avoidance}). The detection model and tracker design parameters are defined in Sections~\ref{subsec:OD} and~\ref{subsec:OT}.

We investigate detection and trigger performance to assess the effectiveness of the perception and avoidance system. Additionally, we conduct flight experiments to assess the success of avoidance on the following aspects. \textit{Detection}: investigate ball detection within a distance bin from 0-5\,m. \textit{Trigger}: investigate an avoidance trigger for thrown balls (we consider a thrown ball not issuing a trigger as failure). \textit{Flight experiments}: validate the physical maneuver of the quadrotor to avoid the thrown ball. 
Every throw is aimed at the UAV, and no avoidance was triggered without a thrown ball (i.e., no false triggers). The stationary experiment evaluates only detection and command generation, since the UAV does not execute the maneuver.

\subsection{Stationary Experiment}
\label{subsec:statexp}

\begin{figure}
  \centering
  \includegraphics[width=\linewidth]{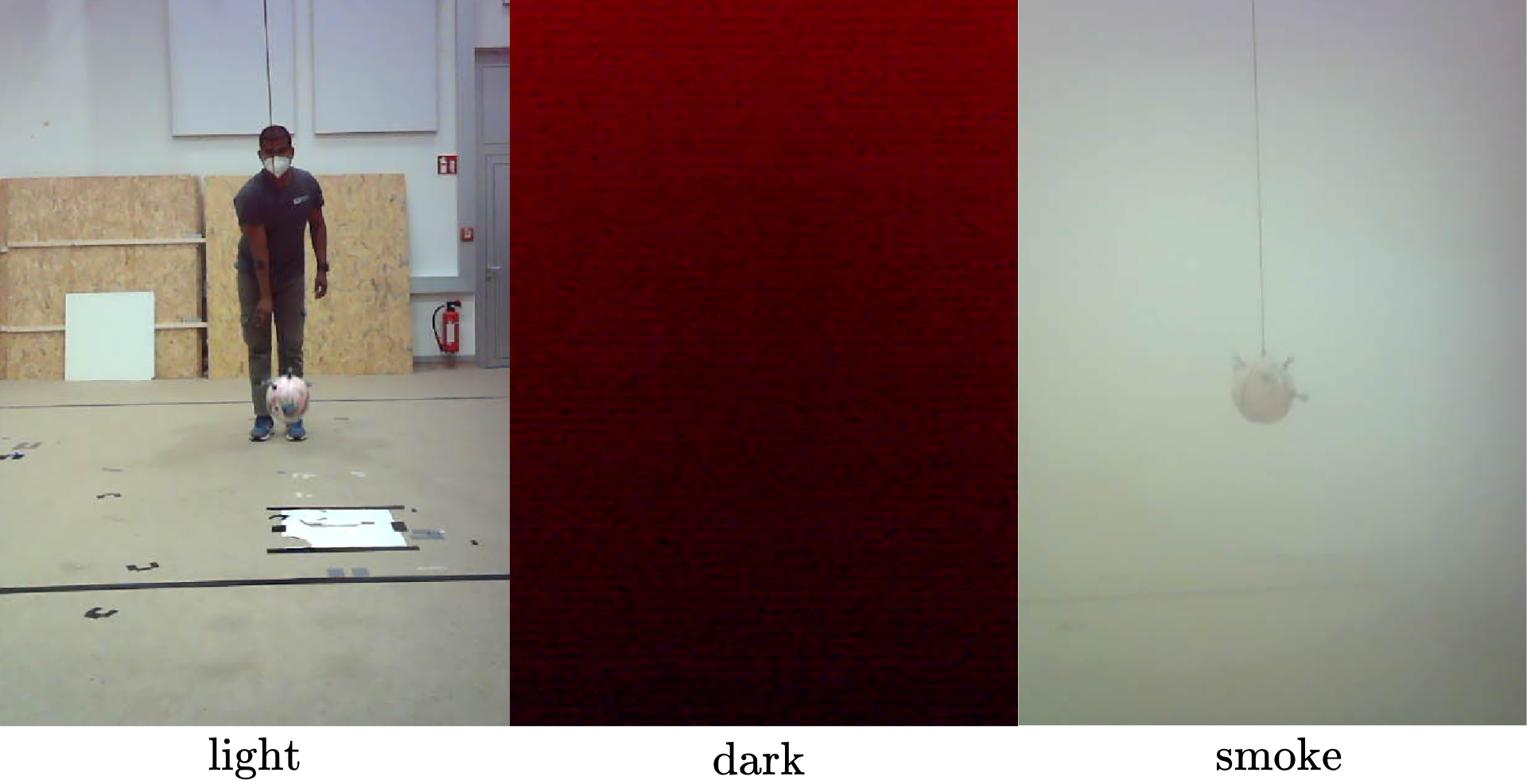}
  \caption{Three scene conditions in the HiL experiment for handball, where the camera fails to capture the ball in darkness and smoke.}
  \Description{Capture instance of stationary experiment in three conditions (light, dark, smoke) inside a drone hall from onboard camera on the quadrotor.}
  \label{fig:scenes}
\end{figure}

We assess the accuracy and trigger performance of the proposed system using ground truth from the MOCAP system. During the experiment, the UAV is kept stationary, with perception via RADAR and control via PX4, running in a loop on a laptop without a GPU. The avoidance commands are computed and issued, but not executed. The throws are performed for 3 different sizes of ball ($\leq$0.1\,m tennis ball, $\leq$0.2\,m handball, and $\leq$0.3\,m basketball) in three conditions: light, dark, and smoke, as shown in Figure~\ref{fig:scenes}. The ball was swung with the help of a stand. Each ball was thrown 50 times for the dark and light scenes, and we recorded a ROS bag containing RADAR perception data, the camera stream, and MOCAP ground truth. For the smoke scene, we recorded 30 throws without the MOCAP ground truth, as it failed in the smoke.

The detection model output state is matched to the temporally nearest MOCAP ball pose using a nearest-neighbor join: for every detection state timestamp $t_d$, the aim is to find the ball frame $t_b$ minimizing $|t_d - t_b|$. Only pairs satisfying $|t_d - t_b| \leq 20\,\text{ms}$ are retained. RADAR coordinates are sign-inverted during preprocessing to align with the MOCAP reference frame. An independently measured rigid-body offset between the RADAR antenna phase center and the MOCAP origin is compensated on every axis, i.e., $\delta_x = 17.67\,\text{mm}, \delta_y = 239.25\,\text{mm}, \delta_z = 4.83\,\text{mm}$. For the balls, the RADAR returns from the bottom (nearest) surface of the ball, while the MOCAP system reports the position of the reflective marker cap mounted on the top of the ball (see Figure~\ref{fig:scenes}). Therefore, the $z$ axis is adjusted for a full ball diameter plus the marker cap radius, using the corresponding radius of each ball. For each ball type, corrected 3D error norms from all throws were pooled, and samples with an absolute modified Z-score greater than 3.5 were removed before computing the mean, standard deviation (std), median, and median absolute deviation (MAD).

\subsubsection{Obstacle Detection}
\label{subsubsec:detection}

Table~\ref{tab:det_accuracy} presents the mean, MAD, and std of position error for three conditions for all ball sizes together, partitioned into four distance bins from 0 to 5\,m (MOCAP distance). $n$ represents the total number of detections in the respective distance bin across the $N$ throws for each condition. For the light and dark conditions, the $x$ mean error is at most 0.15\,m, with std and MAD below 0.25\,m  and 0.05\,m, respectively. The maximum y-axis mean, MAD, and std are 0.93\,m, 0.38\,m, and 0.58\,m, respectively, while the corresponding maximum $z$-axis values are 0.87\,m, 0.43\,m, and 0.57\,m. The larger $y$- and $z$-axis errors are caused by the combined effects of the $\approx$0.23\,m range resolution, sparse returns from the low-RCS non-metallic balls, and artifacts of temporal clustering, which is also reflected in the work of~\cite{pursueradar2020} and~\cite{batmobility2023}. The $y$ and $z$ mean errors ($\leq$0.93\,m and 0.87\,m) are of the order of $R_{sum} = 0.8$\,m, so safety relies on the margins of the controller. The controller therefore keeps an extra lateral clearance $d_{cl}$ and a time margin $\Delta$ (see Eq.~\eqref{eq:acbf}). The outlier filter removes mostly ghost detections due to multipath, resulting in the mean error over all balls and distance bins of $(0.12, 0.64, 0.74)$\,m in light and $(0.12, 0.68, 0.65)$\,m in dark.

\begin{table}
\centering
\caption{Error metrics (Light/Dark) computed against MOCAP ground truth after outlier removal.}
\label{tab:det_accuracy}
\setlength{\tabcolsep}{2pt}
\renewcommand{\arraystretch}{1.1}
\begin{tabular}{cccccc}
\hline
\textbf{Cond.} & \textbf{Dist.} & \textbf{$n$} &\textbf{Mean(m)} & \textbf{MAD(m)} & \textbf{Std(m)}\\
\textbf{} & \textbf{} & \textbf{} & \textbf{$(x,y,z)$} & \textbf{$(x,y,z)$} & \textbf{$(x,y,z)$}\\
\hline
\multirow{4}{*}{\begin{tabular}{c} Light \\ ($N$=150) \end{tabular}} & 0-1 & 116 & (0.15,0.72,0.40) & (0.04,0.26,0.17) & (0.24,0.48,0.32)\\
& 1-2 & 260 & (0.15,0.83,0.69) & (0.05,0.30,0.36) & (0.20,0.23,0.46)\\
& 2-3 & 336 & (0.13,0.85,0.87) & (0.05,0.36,0.40) & (0.19,0.23,0.53)\\
& 3-5 & 684 & (0.10,0.46,0.75) & (0.05,0.21,0.43) & (0.08,0.38,0.55)\\
\hline
\multirow{4}{*}{\begin{tabular}{c} Dark \\ ($N$=150) \end{tabular}} & 0-1 & 339 & (0.14,0.81,0.42) & (0.03,0.38,0.22) & (0.25,0.58,0.30)\\
& 1-2 & 817 & (0.14,0.93,0.57) & (0.04,0.25,0.27) & (0.20,0.58,0.38)\\
& 2-3 & 1072 & (0.12,0.77,0.69) & (0.04,0.34,0.37) & (0.10,0.49,0.49)\\
& 3-5 & 1811 & (0.11,0.49,0.71) & (0.04,0.22,0.37) & (0.07,0.37,0.57)\\
\hline
\end{tabular}%
\end{table}

\begin{table}
\centering
\caption{Dispersion of the detected positions without ground truth, computed identically for all conditions.}
\label{tab:dispersion}
\setlength{\tabcolsep}{2pt}
\renewcommand{\arraystretch}{1.1}
\begin{tabular}{ccccc}
\hline
\textbf{Cond.} & \textbf{Dist.} & \textbf{$n$} & \textbf{MAD(m)} & \textbf{Std(m)}\\
 & & & \textbf{$(x,y,z)$} & \textbf{$(x,y,z)$}\\
\hline
\multirow{4}{*}{\begin{tabular}{c} Light \\ ($N$=150) \end{tabular}} & 0-1 & 119 & (0.01,0.03,0.05) & (0.04,0.06,0.12)\\
& 1-2 & 803 & (0.03,0.11,0.11) & (0.10,0.21,0.28)\\
& 2-3 & 736 & (0.02,0.12,0.10) & (0.05,0.22,0.22)\\
& 3-5 & 2379 & (0.01,0.25,0.32) & (0.04,0.43,0.57)\\
\hline
\multirow{4}{*}{\begin{tabular}{c} Dark \\ ($N$=150) \end{tabular}} & 0-1 & 175 & (0.01,0.02,0.04) & (0.03,0.04,0.08)\\
& 1-2 & 769 & (0.01,0.09,0.09) & (0.08,0.19,0.25)\\
& 2-3 & 813 & (0.01,0.13,0.10) & (0.03,0.21,0.24)\\
& 3-5 & 2760 & (0.02,0.28,0.34) & (0.05,0.48,0.63)\\
\hline
\multirow{4}{*}{\begin{tabular}{c} Smoke \\ ($N$=90) \end{tabular}} & 0-1 & 42 & (0.01,0.02,0.05) & (0.03,0.03,0.09)\\
& 1-2 & 179 & (0.03,0.08,0.06) & (0.08,0.15,0.18)\\
& 2-3 & 242 & (0.00,0.09,0.07) & (0.02,0.15,0.17)\\
& 3-5 & 396 & (0.01,0.14,0.14) & (0.03,0.32,0.35)\\
\hline
\end{tabular}
\end{table}

Since MOCAP was not available in smoke, accuracy can not be reported. However, all three conditions use the same pendulum setup, release point, and RADAR configuration, so the dispersion of the detected positions within a distance bin can be compared across different lighting conditions. Table~\ref{tab:dispersion} lists the dispersion of the detected positions for all three conditions, computed identically from the detections (RADAR-detected distance). In all distance bins, the smoke dispersion (std up to $(0.08, 0.32, 0.35)$\,m) is lower than in light (std up to $(0.10, 0.43, 0.57)$\,m) and in dark (std up to $(0.08, 0.48, 0.63)$\,m). Moreover, the mean detected position per distance bin in smoke deviates by at most 0.17\,m ($x$, $y$) and 0.23\,m ($z$) from light and dark, i.e., less than the detection error itself, which indicates no smoke-specific bias.

\begin{table}
\centering
\caption{Detection of proposed mmWave RADAR sensing in different distance bins and avoidance trigger of DR-CVaR on the stationary UAV for each ball.}
\label{tab:det_rate}
\resizebox{\columnwidth}{!}{%
\begin{tabular}{cccccccc}
\hline
\textbf{Cond.} & \textbf{Dist.} & \multicolumn{3}{c}{\textbf{Detection Rate}} & \multicolumn{3}{c}{\textbf{Avoidance Trigger Rate}} \\ \cline{3-8}
 &  & \textbf{$\leq$0.1} & \textbf{$\leq$0.2} & \textbf{$\leq$0.3} & \textbf{$\leq$0.1} & \textbf{$\leq$0.2} & \textbf{$\leq$0.3} \\
\hline
\multirow{4}{*}{\begin{tabular}{c} Light \\ ($N$=50) \end{tabular}} & 0-1 & 50.0 & 42.0 & 44.0 & \multirow{4}{*}{\begin{tabular}{c} 94.0 \end{tabular}} & \multirow{4}{*}{\begin{tabular}{c} 92.0 \end{tabular}} & \multirow{4}{*}{\begin{tabular}{c} 92.0 \end{tabular}}\\
& 1-2 & 78.0 & 84.0 & 64.0 \\
& 2-3 & 84.0 & 96.0 & 86.0 \\
& 3-5 & 96.0 & 98.0 & 96.0 \\
\hline
\multirow{4}{*}{\begin{tabular}{c} Dark \\ ($N$=50) \end{tabular}} & 0-1 & 48.0 & 34.0 & 46.0 & \multirow{4}{*}{\begin{tabular}{c} 88.0 \end{tabular}} & \multirow{4}{*}{\begin{tabular}{c} 92.0 \end{tabular}} & \multirow{4}{*}{\begin{tabular}{c} 88.0 \end{tabular}} \\
& 1-2 & 86.0 & 80.0 & 88.0 \\
& 2-3 & 80.0 & 98.0 & 98.0 \\
& 3-5 & 94.0 & 98.0 & 100.0 \\
\hline
\multirow{4}{*}{\begin{tabular}{c} Smoke \\ ($N$=30) \end{tabular}} & 0-1 & 6.7 & 10.0 & 53.3 & \multirow{4}{*}{\begin{tabular}{c} 70.0 \end{tabular}} & \multirow{4}{*}{\begin{tabular}{c} 86.7 \end{tabular}} & \multirow{4}{*}{\begin{tabular}{c} 96.7 \end{tabular}} \\
& 1-2 & 60.0 & 86.7 & 80.0 & \\
& 2-3 & 66.7 & 76.7 & 93.3 & \\
& 3-5 & 76.7 & 80.0 & 100.0 & \\
\hline
\end{tabular}%
}%
\end{table}

Table~\ref{tab:det_rate} shows the detection rate of mmWave RADAR perception and the DR-CVaR trigger rate of the DR-ACBF avoidance pipeline. The detection rate is defined as the number of instances of balls detected in each bin divided by the total number of throws. Similarly, the avoidance trigger rate is the fraction of throws for which the avoidance was triggered once without any false triggers. Detection is lowest in the 0-1\,m bin, due to the throw dynamics and the range resolution, and for the smallest non-metallic ball (RCS 0.001\,m$^2$), whose avoidance trigger rate in smoke is 70\%. This is also evident in Table~\ref{tab:det_accuracy}. However, the detection rate for the tennis ball in Table~\ref{tab:det_rate} has increased for distances beyond 3\,m, which is also beneficial for fast DOA. Averaged over bins beyond 1\,m, the detection rates are 87\% in light, 91\% in dark, and 80\% in smoke. MOCAP confirms that all 300 light and dark throws passed within $R_{sum}$ of the RADAR, and no trigger occurred while the ball was at rest before its release, i.e., there were no unnecessary triggers. More importantly, the camera stream fails to capture the ball in darkness and smoke (see Figure~\ref{fig:scenes}).

\subsubsection{Avoidance Trigger Performance}
\label{subsubsec:avoidance}

This section validates the performance of the DR-ACBF avoidance framework for the mmWave RADAR perception. We evaluate the performance based on the avoidance trigger distance $d_{t}$, i.e., the tracked obstacle distance at the trigger time. In light and dark, we also report the miss distance $b$, i.e., the closest approach of the ball center to the RADAR measured by MOCAP. In this case, it equals the miss distance without avoidance since the UAV is stationary.
In addition to the module-level execution latencies (i.e., $t_{end}$), Table~\ref{tab:avoidance_stat} reports three logged timing metrics: the decision latency $t_{dec}$ from the first track of the obstacle to the trigger, the command time $t_{cm}$ from the trigger to the first logged avoidance command, and $t_{ca}$, the predicted time of the worst risk at the trigger. No vehicle motion is involved in any of these metrics. The row $d_t^{\min}$ gives the minimum over all throws of a ball. Since the ball speed varies along the throw, we report its average and maximum instead of its mean and std.

\begin{table}
\centering
\caption{DR-ACBF trigger and timing metrics on the stationary UAV for the throws in Table~\ref{tab:det_rate}.}
\label{tab:avoidance_stat}
\setlength{\tabcolsep}{4pt} 
\renewcommand{\arraystretch}{1.1}
\begin{tabular}{cccccccc} 
\hline
\textbf{Cond.} & \textbf{Metric} & \multicolumn{2}{c}{\textbf{$\leq$0.1}} & \multicolumn{2}{c}{\textbf{$\leq$0.2}}  & \multicolumn{2}{c}{\textbf{$\leq$0.3}} \\ \cline{3-8}
 & & \textbf{Mean} & \textbf{Std.} & \textbf{Mean} & \textbf{Std.} & \textbf{Mean} & \textbf{Std.} \\
\hline
\multirow{6}{*}{\begin{tabular}{c} Light \end{tabular}} & $d_{t}$ (m) & 2.91 & 0.72 & 3.46 & 0.62 & 3.71 & 0.56 \\
& $d_{t}^{\min}$ (m) & 1.27 & -- & 1.78 & -- & 1.64 & -- \\
& $t_{dec}$ (s) & 0.06 & 0.08 & 0.06 & 0.07 & 0.02 & 0.03 \\
& $t_{cm}$ (s) & 0.01 & 0.00 & 0.00 & 0.00 & 0.01 & 0.01 \\
& $t_{ca}$ (s) & 1.12 & 0.79 & 1.23 & 0.74 & 0.98 & 0.64 \\
& $b$ (m) & 0.69 & 0.02 & 0.69 & 0.01 & 0.70 & 0.01 \\
\hline
\multirow{6}{*}{\begin{tabular}{c} Dark \end{tabular}} &  $d_{t}$ (m) & 2.73 & 0.58 & 3.20 & 0.80 & 4.03 & 0.40 \\
& $d_{t}^{\min}$ (m) & 1.55 & -- & 1.22 & -- & 2.36 & -- \\
& $t_{dec}$ (s) & 0.08 & 0.11 & 0.06 & 0.07 & 0.06 & 0.09 \\
& $t_{cm}$ (s) & 0.01 & 0.01 & 0.01 & 0.00 & 0.01 & 0.01 \\
& $t_{ca}$ (s) & 1.21 & 0.84 & 1.31 & 0.75 & 1.02 & 0.60 \\
& $b$ (m) & 0.68 & 0.03 & 0.68 & 0.03 & 0.70 & 0.01 \\
\hline
\multirow{5}{*}{\begin{tabular}{c} Smoke \end{tabular}} &  $d_{t}$ (m) & 2.12 & 0.54 & 2.23 & 0.76 & 2.98 & 1.02 \\
& $d_{t}^{\min}$ (m) & 1.22 & -- & 1.37 & -- & 0.60 & -- \\
& $t_{dec}$ (s) & 0.04 & 0.09 & 0.06 & 0.08 & 0.02 & 0.03 \\
& $t_{cm}$ (s) & 0.01 & 0.01 & 0.01 & 0.01 & 0.01 & 0.01 \\
& $t_{ca}$ (s) & 1.74 & 0.60 & 1.65 & 0.62 & 0.89 & 0.74 \\
\hline
\end{tabular}
\end{table}

The avoidance trigger was successful in 89\% at an average ball velocity of over 3.1\,m/s and a maximum of 8.40\,m/s. The group-mean $d_t$ ranged from 2.12\,m to 4.03\,m, but individual throws triggered as late as 1.22\,m in light and dark and 0.60\,m in smoke. MOCAP shows that all light and dark throws were nearly head-on, with a miss distance $b$ of 0.49--0.74\,m $< R_{sum}$, i.e., every throw would have hit the UAV without avoidance. The means of $t_{dec}$ and $t_{cm}$ are at most 0.08\,s and 0.01\,s, with a maximum $t_{cm}$ of 0.055\,s, for objects thrown from 5\,m with a maximum mean $t_{ca}$ of 1.74\,s.
In all conditions, larger balls are detected and triggered earlier: the $\leq$0.3 ball triggered at a mean $d_t$ of 3.71\,m, 4.03\,m, and 2.98\,m in light, dark, and smoke, compared to 2.91\,m, 2.73\,m, and 2.12\,m for the $\leq$0.1 ball.

\subsection{Indoor Flight Experiment}
\label{subsec:flyexp}

The small-scale UAV is limited in agility by its total size and is flown indoors with the MOCAP position system at a height of $\approx$1.5\,m. For the flight experiment, the UAV was tethered with a rope, which was handled by a human to ensure safety. Only the basketball was used because it had the highest avoidance trigger rate; it was tied to the rope and then thrown towards the UAVs. The smoke scene was excluded due to MOCAP failure. 

\begin{table}
  \centering
  \caption{Indoor flights with a dynamic obstacle for hovering and moving forward at 1\,m/s conditions (5 flights each).}
  \label{tab:combined_flight_metrics}
  \begin{tabular}{cccccccc}
    \hline
    \multirow{2}{*}{\textbf{Cond.}} & \multirow{2}{*}{\textbf{}} & \multicolumn{3}{c}{\textbf{Hovering}} & \multicolumn{3}{c}{\textbf{Moving Forward}} \\ \cline{3-8}
    & & $t_{dec}$ & $t_{cm}$ & $d_t$ & $t_{dec}$ & $t_{cm}$ & $d_t$ \\
    & & (s) & (s) & (m) & (s) & (s) & (m) \\
    \hline
    \multirow{3}{*}{Light} & mean & 0.10 & 0.10 & 3.48 & 0.06 & 0.15 & 3.94 \\
                           & std  & 0.04 & 0.13 & 1.37 & 0.02 & 0.08 & 1.55 \\
                           & worst & 0.14 & 0.33 & 1.37 & 0.09 & 0.22 & 2.40 \\
    \hline
    \multirow{3}{*}{Dark}  & mean & 0.10 & 0.02 & 3.02 & 0.06 & 0.23 & 2.94 \\
                           & std  & 0.07 & 0.01 & 1.47 & 0.03 & 0.23 & 0.82 \\
                           & worst & 0.22 & 0.03 & 1.14 & 0.11 & 0.48 & 1.53 \\
    \hline
  \end{tabular}
\end{table}

Table~\ref{tab:combined_flight_metrics} reports hovering and forward flights in light and dark. For all 20 flights, the average speed of the ball was 6.43\,m/s, with a maximum of 6.8\,m/s.
The rows labeled "worst" show the maximum of the times and the minimum of the distances over the 5 flights.
For hovering, both conditions show similar decision latency ($t_{dec} = 0.10 \pm 0.04$\,s in light and $0.10 \pm 0.07$\,s in dark). The $d_t$ in light and dark shows a small difference, from $3.48 \pm 1.37$\,m in light to $3.02 \pm 1.47$\,m in dark. 
The command time $t_{cm}$ is lower in dark conditions ($0.02 \pm 0.01$\,s) than in light conditions ($0.10 \pm 0.13$\,s). The $t_{cm}$ measures when DR-ACBF acceleration feedforward along the dodge axis becomes non-zero, which depends on the encounter geometry.
In the forward flights, $t_{dec}$  improves compared to hovering, decreasing to $0.06 \pm 0.02$\,s in light and $0.06 \pm 0.03$\,s in dark.
However, $t_{cm}$ increases, particularly in dark conditions ($0.23 \pm 0.23$\,s), since $t_{cm}$ depends on the encounter geometry. The $d_t$ is $3.94 \pm 1.55$\,m in light and $2.94 \pm 0.82$\,m in dark, respectively. All 20 flights were successful avoidance maneuvers without contact. The safety tether did not intervene in any of the 20 avoidance maneuvers.

\subsection{Outdoor Flight Experiment}
\label{subsec:outdoor}

We validated our system in an untethered outdoor setting without MOCAP. The UAV estimated its state using GPS for position and a range finder for altitude. The quadrotor was flown in daylight while the three balls were thrown towards it. For hovering, we conducted five flights, each including all three balls, i.e., 5 avoidance maneuvers per ball. In forward flight along a 14\,m straight line at a nominal speed of 3.5\,m/s, we performed 3 flights for each ball. Since the PX4 also corrects the lag behind the moving position setpoint, the measured speed at the trigger was 4.4--4.7\,m/s, respectively, and this measured speed enters the closing speed. In all flights, the avoidance was successful and completed, i.e., the maneuver was released after the ball had passed or left the FoV. In 8 further flights, balls that passed at least 1\,m from the UAV or moved away from it were tracked, and no avoidance was triggered.

Table~\ref{tab:outdoor} reports the performance metrics from the RADAR track for hovering and forward flight at 3.5\,m/s for the outdoor flight experiment. In the indoor experiment, the logged avoidance command was the DR-ACBF acceleration, which became nonzero only when the ball was close enough to activate Eq.~\eqref{eq:acbf}. Indoors, the ball was thrown away from 5\,m, against $\geq$10\,m outdoors.  For most outdoor experiments, the avoidance command sent to the PX4 is also logged, which follows within one or two control cycles. Hence, the smaller $t_{cm}$ outdoors.
In hovering,  the tennis ball was recorded as the fastest throw (10.3--13.8\,m/s), triggered earliest, at $9.54 \pm 1.80$\,m. The handball triggered at $8.26 \pm 2.44$\,m and the basketball at $8.17 \pm 2.90$\,m, i.e., more than twice as far as indoors (Table~\ref{tab:combined_flight_metrics}). The faster outdoor throws were tracked over a longer distance and thus triggered earlier. 
$t_{dec}$ was at most 0.05\,s. $t_{cm}$ was 21--32\,ms in all 15 avoidance. 
In forward flight at 3.5\,m/s, the UAV speed adds to the closing speed (12.2--14.6\,m/s), and the balls are triggered at 3.5--10.0\,m, closer than in hovering. $t_{dec}$ was at most 0.08\,s except for one throw (0.22\,s), and $t_{cm}$ was 20--40\,ms.

\begin{table}
\centering

\caption{Outdoor flights for hovering (5 avoidance per ball) and forward flight at 3.5\,m/s (3 avoidance per ball).}
\label{tab:outdoor}
\setlength{\tabcolsep}{3pt}
\renewcommand{\arraystretch}{1.1}
\begin{tabular}{cccccccc}
\hline
\multirow{2}{*}{\textbf{Ball}} & & \multicolumn{3}{c}{\textbf{Hovering}} & \multicolumn{3}{c}{\textbf{Moving Forward}} \\ \cline{3-8}
 & & $t_{dec}$ & $t_{cm}$ & $d_t$ & $t_{dec}$ & $t_{cm}$ & $d_t$ \\
 & & (s) & (s) & (m) & (s) & (s) & (m) \\
\hline
\multirow{3}{*}{Tennis} & mean & 0.03 & 0.03 & 9.54 & 0.05 & 0.03 & 5.67 \\
 & std & 0.00 & 0.00 & 1.80 & 0.03 & 0.01 & 0.45 \\
 & worst & 0.03 & 0.03 & 8.01 & 0.08 & 0.04 & 5.15 \\
\hline
\multirow{3}{*}{Handball} & mean & 0.03 & 0.03 & 8.26 & 0.09 & 0.03 & 7.99 \\
 & std & 0.01 & 0.00 & 2.44 & 0.11 & 0.01 & 2.79 \\
 & worst & 0.05 & 0.03 & 5.72 & 0.22 & 0.03 & 4.81 \\
\hline
\multirow{3}{*}{Basketball} & mean & 0.03 & 0.02 & 8.17 & 0.04 & 0.04 & 6.84 \\
 & std & 0.01 & 0.00 & 2.90 & 0.01 & 0.01 & 3.08 \\
 & worst & 0.05 & 0.03 & 4.23 & 0.05 & 0.04 & 3.53 \\
\hline
\end{tabular}
\end{table}

We also performed forward flights at 2 and 5\,m/s with measured speeds at the trigger of 1.3--2.5 and 5.9--6.6\,m/s, respectively. Flying forward adds the UAV speed to the closing speed $v_c$. For the tennis ball, $v_c$ rises from 11.4\,m/s in hovering to 13.4 and 14.6\,m/s at 2 and 3.5\,m/s, and $d_{crit}$ grows from 5.0 to 6.1 and 6.3\,m. The trigger distance does not increase with $d_{crit}$, as $d_t$ is limited by the ball's detection range. As Eq.~\eqref{eq:vego} predicts, $d_t > d_{crit}$ therefore held in all hovering avoidance, but failed in one throw at each forward speed. At 2\,m/s, the handball triggered at 5.27\,m with $d_{crit}=5.30$\,m. At 3.5 and 5\,m/s, the tennis ball triggered at 5.95\,m and 4.46\,m with $d_{crit}=6.33$\,m and 4.85\,m. For these throws, the condition $d_t > d_{crit}$ did not hold, but avoidance was successful because $d_t > d_{crit}$ is a sufficient condition. The $d_t > d_{crit}$ condition is also conservative since it assumes a constant closing speed and head-on throw ($b=0$, Eq.~\eqref{eq:tevade}), whereas the dodge also brakes the UAV.

\subsection{Computation Load}
\label{subsec:comp}

Table~\ref{tab:latencyandload} lists the runtime of the seven components on the laptop and the Pi4B. Clustering represents the RADAR pointcloud acquisition and the execution time of DBSCAN applied to the raw mmWave RADAR pointcloud. In Table~\ref{tab:latencyandload}, the other rows are the remaining detection steps ($t_{det}$) and the avoidance step ($t_{avoid}$). The execution time per frame ($t_{end}$) almost doubles from 7.63\,ms on the laptop to 13.65\,ms on the Pi4B. On the laptop, clustering dominates (78.2\%), since it accounts for the data stream delay. On the Pi4B, it drops to 44.0\%, since the tracker (22.9\%) and the candidate update (18.3\%) become more expensive, and the ACBF (0.76\,ms) shares the processor with the flight control. The RADAR delivers a frame every 10\,ms (100.0\,Hz on average over all flights), each frame is processed within $t_{end}$, and the control loop sends the latest command to the PX4 at 200\,Hz. On the Pi4B, $t_{end}$ exceeds the 10\,ms frame period, but the tracker always processes the newest frame (Section~\ref{subsec:stream}), so the delay does not add up. The avoidance command follows the first frame after $t_{lat} = t_{end} + t_{dec}$ (Figure~\ref{fig:sys}), with a median of 75\,ms indoors (20 flights) and 48\,ms outdoors (24 avoidance), dominated by the decision time $t_{dec}$ (median 61\,ms indoors).

We compare this with the event camera DOA in~\cite{doafalanga2020}. The event camera outputs individual events continuously, and~\cite{doafalanga2020} collects them in a 10\,ms time window. Processing one window takes 3.5\,ms on a Jetson TX2 GPU, since the events must first be compensated for camera rotation, thresholded, filtered, and clustered to detect the obstacle; the position estimation, the Kalman filter, and the avoidance command are not included in this 3.5\,ms. The event camera is therefore about four times faster per window. In practice, this gain is small; the difference of about 10\,ms increases $d_{crit}$ by only 0.15\,m at our maximum closing speed of 14.6\,m/s (Eq.~\eqref{eq:critical_distance}). The detection range matters much more. The event camera dodges obstacles from about 3\,m, whereas our trigger fired at 8.2-9.5\,m on average in outdoor hovering (Table~\ref{tab:outdoor}). The RADAR also measures velocity from the first frame, and our tracking compensates for the UAV's full motion, not just its rotation. Thus, despite its higher latency compared to an event camera, our system detects obstacles earlier and leaves more time to evade. This suggests that detection range matters more than latency for fast DOA.

\begin{table}
\centering
\small 
\setlength{\tabcolsep}{2pt} 
\caption{Latency and computational load of our mmWave RADAR-based DOA framework.}
\label{tab:latencyandload}
\renewcommand{\arraystretch}{1.1}
\begin{tabular*}{\columnwidth}{@{\extracolsep{\fill}}llcccccc}
\hline
& \multirow{2}{*}{\textbf{Comp.}} & \multicolumn{3}{c}{\textbf{Laptop ($N$=390)}} & \multicolumn{3}{c}{\textbf{Pi4B ($N$=20)}} \\ \cline{3-8} 
& & \textbf{M} & \textbf{S} & \textbf{L} & \textbf{M} & \textbf{S} & \textbf{L} \\ \hline
\multirow{4}{*}{$t_{det}$} & Clustering & 5.97 & 15.2 & 78.2 & 6.00 & 2.9 & 44.0 \\
& Candidates & 0.38 & 0.7 & 4.9 & 2.50 & 1.5 & 18.3 \\
& Boxes      & 0.08 & 0.4 & 1.0 & 0.71 & 1.4 & 5.2 \\
& Tracker    & 0.58 & 3.6 & 7.6 & 3.13 & 7.8 & 22.9 \\ \hline
\multirow{3}{*}{$t_{avoid}$} & Trigger    & 0.26 & 1.4 & 3.4 & 0.51 & 5.6 & 3.7 \\
& ACBF       & 0.30 & 1.6 & 3.9 & 0.76 & 3.6 & 5.6 \\
& Projection & 0.06 & 0.5 & 0.8 & 0.04 & 0.3 & 0.3 \\ \hline
$t_{end}$ & & \textbf{7.63} & - & \textbf{100} & \textbf{13.65} & - & \textbf{100} \\ \hline
\end{tabular*}
\raggedright \scriptsize M: Mean (ms), S: Std (ms), L: Load (\%)
\end{table}

\subsection{Ablation}
\label{subsec:ablation}

We first ablate the RADAR rate and the uncertainty terms. We then analyze the sensitivity to the acceleration budget.

\subsubsection{RADAR sampling rate:}

We replay the recorded RADAR frames of the stationary throws at 100\,Hz, 50\,Hz, and 33\,Hz, using every frame, every second frame, or every third frame. For each throw, we check $d_t > d_{crit}$ with the MOCAP distance and closing speed at the trigger (Table~\ref{tab:ablation}). At 33\,Hz, the trigger fires later, with a median $d_t$ of 3.21 instead of 3.74\,m. The throws with $d_t > d_{crit}$ drop from 283 to 183 out of 300, mostly for the small tennis ball (90 to 36 out of 100). The drop is smaller when $d_{crit}$ is based on the measured miss distance $b$ (from 298 to 288). In smoke, where we use the tracked distance and speed, it drops from 69 out of 90 to 59.

\begin{table}
\centering
\caption{Effect of the RADAR rate on the trigger (median $t_{dec}$ and $d_t$).}
\label{tab:ablation}
\setlength{\tabcolsep}{3pt}
\renewcommand{\arraystretch}{1.1}
\begin{tabular}{cccccc}
\hline
\textbf{Rate} & \textbf{$d_t$} & \textbf{$t_{dec}$} & \multicolumn{3}{c}{\textbf{Throws with $d_t > d_{crit}$}} \\ \cline{4-6}
(Hz) & (m) & (ms) & \multicolumn{2}{c}{Light and dark} & Smoke \\
 & & & head-on & measured $b$ & head-on \\
\hline
100 & 3.74 & 3 & 283 & 298 & 69 \\
50 & 3.53 & 28 & 258 & 297 & 67 \\
33 & 3.21 & 31 & 183 & 288 & 59 \\
\hline
\end{tabular}
\end{table}

\subsubsection{Uncertainty Terms:}

The DR-CVaR trigger and the DR-ACBF can include the covariance of the tracks. The DR-ACBF can also include a closing-speed uncertainty $\sigma_v$. We first test the covariance in the trigger on the replay at 100\,Hz (see Table~\ref{tab:ablation_cov}). There is no change in light and dark, because the trigger already fires at the first track. In smoke, the detections are noisiest. Therefore, in smoke, the covariance increases the number of throws with $d_t > d_{crit}$ to 79 of 90 (without covariance:  69 of 90) and no trigger while the ball is at rest. We then performed three hovering flights with one throw for each ball, under three conditions: (i) with the covariance, (ii) with $\sigma_v = 0.3$\,m/s, and (iii) with both. We compare these flights with the 15 hovering avoidance flights in Table~\ref{tab:outdoor}, flown without the covariance and $\sigma_v$. All 9 new avoidance (3 balls and 3 conditions) were triggered in time and satisfied $d_t > d_{crit}$ for $b = 0$. The difference lies in the avoidance-dodge maneuver. With the covariance, the DR-ACBF pushed the acceleration to the dodge in 5 of 6 avoidance attempts within 0.33\,s after the trigger. Without covariance, this happened in only 2 of 15 avoidance attempts, after 0.73\,s or later. With $\sigma_v$ alone, the DR-ACBF never added acceleration to the dodge.

\begin{table}
\centering
\caption{Effect of the track covariance in the trigger (at 100\,Hz; $b=0$).}
\label{tab:ablation_cov}
\setlength{\tabcolsep}{3pt}
\renewcommand{\arraystretch}{1.1}
\begin{tabular}{cccc}
\hline
\textbf{Covariance} & \textbf{Light and dark} & \multicolumn{2}{c}{\textbf{Smoke}} \\ \cline{3-4}
\textbf{in trigger} & $d_t > d_{crit}$ & Triggered & $d_t > d_{crit}$ \\
\hline
Off & 283 & 87 & 69 \\
On & 283 & 89 & 79 \\
\hline
\end{tabular}
\end{table}

\subsubsection{Acceleration budget:}

\begin{table}
\centering
\caption{Sensitivity to acceleration budget on the indoor and outdoor hover/forward avoidance (median $d_{crit}$ (m); $b=0$).}
\label{tab:ablation_accel}
\setlength{\tabcolsep}{1.5pt}
\renewcommand{\arraystretch}{1.0}
\begin{tabular}{cccccccc}
\hline
\textbf{$a$} & \textbf{$t_{evade}$} & \multicolumn{2}{c}{\textbf{Indoor}} & \multicolumn{2}{c}{\textbf{Hover}} & \multicolumn{2}{c}{\textbf{Forward 3.5\,m/s}} \\ \cline{3-8}
(m/s$^2$) & (s) & $d_{crit}$ & $d_t > d_{crit}$ & $d_{crit}$ & $d_t > d_{crit}$ & $d_{crit}$ & $d_t > d_{crit}$ \\
\hline
5 & 0.57 & 4.03 & 5/20 & 6.37 & 14/15 & 9.46 & 2/9 \\
10 & 0.40 & 3.20 & 11/20 & 4.87 & 14/15 & 7.11 & 4/9 \\
15 (ours) & 0.33 & 2.85 & 15/20 & 4.21 & 15/15 & 6.07 & 4/9 \\
25 & 0.25 & 2.48 & 15/20 & 3.54 & 15/15 & 5.02 & 8/9 \\
50 & 0.18 & 2.10 & 16/20 & 2.87 & 15/15 & 3.97 & 8/9 \\
\hline
\end{tabular}
\end{table}

Table~\ref{tab:exp_latency} computes $d_{crit}$ for assumed speed and latencies. Instead of assumed speed and latency, Table~\ref{tab:ablation_accel} uses the measured $v_c$ and $t_{lat}$ of each flight and recomputes $d_{crit}$ with the recorded $d_t$ ($b = 0$, since the miss distance is not measured in flight). Only $t_{evade}$ changes with the acceleration budget $a$, and the trigger itself does not depend on it. A larger acceleration budget helps mainly at high closing speeds. In forward flight at 3.5\,m/s, raising the acceleration budget $a$ to 25\,m/s$^2$ increases the throws with $d_t > d_{crit}$ to 8 of 9 flights, up from 4 of 9 at 15\,m/s$^2$. In hovering, the avoidance condition $d_t > d_{crit}$ was satisfied in all 15 flights for every budget. For indoor flights, an acceleration budget above 15\,m/s$^2$ barely helps; i.e., 15 of 20 flights satisfy the avoidance condition, compared with 16 of 20 flights at 50\,m/s$^2$. This is also because the remaining flights were triggered at 2.4\,m or closer. Therefore, as with shorter latency, a larger acceleration budget cannot compensate for late detection, and the detection range remains the limiting factor.

\subsection{Safety and Failure Analysis}
\label{subsec:failure_analysis}

\subsubsection{Safety Margin Analysis}

We evaluate Eq.~\eqref{eq:critical_distance} for each trial with $R_{sum} = 0.8$\,m, $a = 15$\,m/s$^2$, and a head-on encounter ($b = 0$). We use the closing speed $v_c$ along the LoS at the trigger and the latency $t_{lat} = t_{end} + t_{dec}$ of that trial. In the indoor flights, $d_t > d_{crit}$ held in 15 of the 20 flights (7 of 10 hovering, 8 of 10 forward) at closing speeds of 3.3--6.0\,m/s. In outdoor hovering, all 24 avoidance satisfied $d_t > d_{crit}$, at ball speeds of 8.4--13.8\,m/s and closing speeds of 6.7--13.1\,m/s. In an outdoor forward head-on encounter, $d_t > d_{crit}$ held in 8 of the 15 avoidance cases (Table~\ref{tab:outdoor}). In the replay of the 300 light and dark throws, it held in 283 head-on and in 298 with the miss distance $b$ measured by MOCAP (Table~\ref{tab:ablation}). This is expected, as we evaluate the worst case with $b=0$ and without braking. Nevertheless, the quadrotor avoided the ball in all 20 indoor and all 39 outdoor flights. This shows that the head-on condition is conservative. It is sufficient but not necessary, since it credits neither the braking of the UAV in forward flight nor the lateral offset $b > 0$ of an actual throw.

\subsubsection{Failure Conditions}

Eq.~\eqref{eq:critical_distance} also identifies the three conditions under which clearance is not guaranteed ($d_t \leq d_{crit}$). First, there is \textit{late detection} of small, low-RCS objects such as the tennis ball, which shortens the detection range. For example, a ball detected at 1.5\,m while approaching head-on at 6\,m/s is inside $d_{crit} = 2.84\,$m. The second is a \textit{long decision latency}. In dark hovering, $t_{dec}$ reached 0.22\,s, which increases $d_{crit}$ by $v_c\,t_{dec} = 1.0$\,m at 4.6\,m/s. The third is a \textit{low acceleration budget} of a less agile UAV, which increases $t_{evade}$ (Eq.~\eqref{eq:tevade}). Indoors, $d_t > d_{crit}$ held in 5 of 20 flights at $a = 5$\,m/s$^2$, compared with 15 at 15\,m/s$^2$ (Table~\ref{tab:ablation_accel}). All effects grow in forward flight, where the ego speed increases $v_c$ and thus $d_{crit}$. The 100\,Hz RADAR rate and the track covariance reduce the first two effects (Section~\ref{subsec:ablation}), but they cannot remove them. A faster convergence of the tracker velocity could further shorten $t_{dec}$.

\section{Discussion}
\label{sec:discussion}

Each sensing modality suits a different UAV task, such as LIDAR for mapping and cameras for classification, and our results show that mmWave RADAR can take this role for DOA. mmWave RADAR perception has scaled the most for UAVs, owing to its sensing range of more than 150\,m, compact size, all-weather robustness, and low computational cost. Further, the widespread adoption of mmWave RADAR in automotive applications has significantly reduced its costs. As a general perception modality, however, mmWave RADAR has three limitations. First, the mmWave RADAR (AWR2944PEVM) in theory has a maximum detection range of 250\,m, which is significantly reduced for small-RCS objects~\cite{TI_AWR2944_userguide,TI_RCS_mmWave}. Second, the sparse RADAR pointcloud makes it difficult to estimate the size of objects~\cite{mandaokar_widroit2025}. Third, angular resolution and FoV are poorer than those of cameras or LIDAR~\cite{reviewsensinguav2020}. NN-based detection, fusion, or 4D imaging mmWave RADARs can reduce these limitations~\cite{fusionradar2024}. 

On the Pi4B, the processing takes only 13.65\,ms, which is small compared to the decision latency (median 61\,ms) and the evade time. The detection range is therefore the most important factor for improving RADAR-based DOA. Our experiments used small non-metallic balls, which are a difficult case for RADAR. Moving and metallic obstacles, such as a quadrotor, are easier to detect at long range. The chirp and CFAR configuration can be further optimized for this. Future work will study the effective detection range and track multiple obstacles at longer ranges for multiple-DOA.

\section{Conclusion}
\label{sec:conclusion}
This work presents a lightweight mmWave RADAR-based perception system for real-time DOA on resource-constrained UAVs. Our system achieves robust perception and avoidance at high update rates without relying on GPU or deep learning, enabling practical onboard deployment. Under varying illumination conditions, stationary experiments achieved average detection rates of at least 80\% at distances greater than 1\,m, and detections in smoke were as consistent as in light and dark conditions. In all flight experiments, the UAV avoided the ball in all 20 tethered indoor flights at an average obstacle speed of 6.43\,m/s and in all 39 untethered outdoor flights with three ball sizes, at measured closing speeds up to 14.6\,m/s and measured forward speeds up to 6.6\,m/s. In contrast, balls on a non-collision path did not trigger avoidance. Our mmWave RADAR perception model triggers avoidance 8.2--9.5\,m away, which gives the UAV enough time to evade. A lower latency would add only a small distance to this. Thus, for fast DOA, a long detection range matters more than a low latency. 




\bibliographystyle{SenSys2027/ACM-Reference-Format}
\bibliography{SenSys2027/root}










\end{document}